\documentclass[journal]{IEEEtran}
\usepackage{amsmath,amsfonts}
\usepackage{bm}
\usepackage{algorithmic}
\usepackage{algorithm}
\usepackage{array}
\usepackage[caption=false,font=normalsize,labelfont=sf,textfont=sf]{subfig}
\usepackage{textcomp}
\usepackage{stfloats}
\usepackage{url}
\usepackage{verbatim}
\usepackage{graphicx}
\usepackage{cite}
\usepackage[breaklinks=true]{hyperref}
\usepackage[switch]{lineno}
\usepackage{newtxtext}

\begin{document}

\title{Spatioformer: A Geo-encoded Transformer for Large-Scale Plant Species Richness Prediction}

\author{Yiqing Guo, Karel Mokany, Shaun R. Levick, Jinyan Yang, Peyman Moghadam% <-this % stops a space
\thanks{This study is supported by the Spatiotemporal Activity within CSIRO’s Machine Learning and Artificial Intelligence Future Science Platform, and the Biodiversity Analytics From Space project within CSIRO’s Space Technology Future Science Platform.}
\thanks{The Spatioformer code will be made available at \href{https://github.com/csiro-robotics/Spatioformer}{https://github.com/csiro-robotics/Spatioformer}. The machine learning-ready dataset used in this study for plant species richness prediction in Australia will be released at \href{https://doi.org/10.25919/7d5h-yp05}{https://doi.org/10.25919/7d5h-yp05}.}
\thanks{Yiqing Guo, Karel Mokany, and Jinyan Yang are with CSIRO Environment, Acton, ACT 2601, Australia (emails: yiqing.guo@csiro.au, karel.mokany@csiro.au, jinyan.yang@csiro.au).% <-this % stops a space
}
\thanks{
Shaun R. Levick is with CSIRO Environment, Winnellie, NT 0822, Australia (email: shaun.levick@csiro.au).% <-this % stops a space
}
\thanks{
Peyman Moghadam is with CSIRO Robotics, Data61, Pullenvale, QLD 4069, Australia, and also with the Queensland University of
Technology, Brisbane, QLD 4000, Australia. (email: peyman.moghadam@csiro.au).% <-this % stops a space
}
%\thanks{\textsuperscript{*} The dataset of this study is available at \href{https://data.csiro.au/collection/csiro:62308}{\nolinkurl{https://data.csiro.au/collection/csiro:62308}}.% <-this % stops a space}
}

\markboth{Preprint}%
{Shell \MakeLowercase{\textit{et al.}}: A Sample Article Using IEEEtran.cls for IEEE Journals}


\maketitle

\begin{abstract}
Earth observation data have shown promise in predicting species richness of vascular plants ($\alpha$-diversity), but extending this approach to large spatial scales is challenging because geographically distant regions may exhibit different compositions of plant species ($\beta$-diversity), resulting in a location-dependent relationship between richness and spectral measurements. In order to handle such geolocation dependency, we propose \textit{Spatioformer}, where a novel geolocation encoder is coupled with the transformer model to encode geolocation context into remote sensing imagery. The Spatioformer model compares favourably to state-of-the-art models in richness predictions on a large-scale ground-truth richness dataset (HAVPlot) that consists of 68,170 in-situ richness samples covering diverse landscapes across Australia. The results demonstrate that geolocational information is advantageous in predicting species richness from satellite observations over large spatial scales. With Spatioformer, plant species richness maps over Australia are compiled from Landsat archive for the years from 2015 to 2023. The richness maps produced in this study reveal the spatiotemporal dynamics of plant species richness in Australia, providing supporting evidence to inform effective planning and policy development for plant diversity conservation. Regions of high richness prediction uncertainties are identified, highlighting the need for future in-situ surveys to be conducted in these areas to enhance the prediction accuracy.
\end{abstract}


\section{Introduction}

Australia is home to a large and diverse range of plant species, with over 21,000 known native species of vascular plants and 93\% of these being endemic \cite{chapman2009numbers, chah2022australian}. The richness of plant species, also known as $\alpha$-diversity, is highly important in maintaining the functioning of ecosystems, such as habitat provision, carbon sequestration, and water cycling \cite{feret2012tree, yang2023trajectories, li2023texture, zhang2023morphological}. However, anthropogenic interference, such as deforestation, overgrazing, and urbanisation, has resulted in a decline in plant species richness \cite{ceballos2015accelerated, isbell2023expert}. In response, conservation activities have been initialised and conducted across the country aiming to preserve plant diversity \cite{mokany2020reconciling, shaw2023linking, st2023safeguarding}. Accurate and up-to-date maps of plant species richness will strongly support effective planning and policy-making for these activities \cite{gould2000remote, guo2023plant}. 

Earth observation (EO) data provide rapid and near-real-time estimates of changes in land surface conditions across large regions \cite{madigan2018quantitative, guo2018effective, zhao2024global, guo2019nomination}. This makes remote sensing imagery a favourable data source for plant species richness modelling, as compared with another widely adopted approach where environmental variables, such as temperature, precipitation, soil texture, and topographic heterogeneity, are used as richness predictors \cite{mokany2022patterns}. The reason is that environmental variables drive mainly the environmental potential of plant habitats (\emph{i.e.}, the capacity to sustain a certain level of richness), rather than represent the actual conditions on the ground like those observed by remote sensing satellites. For example, deforestation, floods, and bushfires could cause reduction in richness \cite{gill1999biodiversity}, but such reduction might not be reflected by environmental variables. Therefore, environmental variables are often aimed at predicting the natural patterns in diversity in a pre-intensification reference state, while remote sensing data are more valuable for monitoring actual changes in those patterns.

Australia covers an area of over seven million square kilometres. As a result of the relatively large geographical extents, versatile types of plant habitats are found across the country, differing in their inventories of plant species present that have been shaped by a variety of factors such as biogeographic history, climate, and geography \cite{rosenzweig1995species}. To understand the spatiotemporal distribution of plant species richness, perseverant in-situ field surveys have been conducted over the past several decades. Via various survey campaigns, a wealth of 219,552 richness samples have been gathered across the country as of the year 2022 \cite{mokany2022patterns}. These samples represent a broad range of landscapes across the continent, and therefore present a unique opportunity to unravel the potentially intricate relationship between richness measurements and satellite observations. Nevertheless, geographically distant regions may exhibit distinct assemblages of plant species with differed compositional properties (\emph{i.e.}, $\beta$-diversity \cite{ferrier2007using}), making it challenging to model richness over large spatial scales. Due to spatial variations in plant species composition, a location with a set of plant species would be expected to display quite different spectral features in remote sensing imagery from another location with a dissimilar plant composition, even if the two locations sustain the same richness of species \cite{wang2018influence,mokany2022working}. Through statistical regression analysis for two regions in southeast Australia, previous studies \cite{guo2023plant,guo2022quantitative} suggested that the relationship between plant species richness and hyper/multispectral satellite observations is region-specific. To account for the location dependency, we need a model that is capable of taking in geolocation context when mapping plant species richness over large spatial scales. 

The transformer model, first introduced in \cite{vaswani2017attention}, is built upon the self-attention mechanism \cite{cheng2016long}. The model attends effectively to information of high importance in the input data, as the self-attention module is capable of capturing intricate data structures and dependencies \cite{cheng2016long, xu2022attention}. Initially proposed for language tasks, the transformer model has shown promise for image understanding due to its superior ability over Convolutional Neural Networks (CNNs) in capturing global dependencies between different regions of an image \cite{dosovitskiy2021an}. As a seminal work on applying transformer to image data, the Vision Transformer (ViT) model \cite{dosovitskiy2021an} first divides an image into non-overlapping patches, followed by projecting each patch into a feature vector which is then fed into the self-attention module, with state-of-the-art performance being achieved on benchmark datasets. Given remote sensing imagery captures rich features in the spectral dimension, the SpectralFormer model \cite{hong2021spectralformer} was developed to effectively embed the spectral information. 
%This model was later extended in \cite{mohamed2023factoformer}, where FactoFormer, a factorised transformer, was introduced for the joint learning of spectral and spatial features. 
Recently, FactoFormer~\cite{mohamed2023factoformer} developed a self-supervised factorised spectral--spatial transformer, where attention is computed by individually focusing on spectral and spatial dimensions in each transformer. These studies have demonstrated the effectiveness of transformer in processing remote sensing images (\emph{e.g.}, \cite{mohamed2023factoformer, hong2021spectralformer}), but to advance the model's application to remote sensing images recorded over large spatial scales, geolocational information could be leveraged. Unlike many types of imagery whose semantics are independent of the location where they are recorded, remote sensing images are intrinsically associated with geolocations \cite{ayush2021geography, ge2022geoscience}. Considering that the composition of plant species is location-specific, incorporating geolocation context could be helpful in modelling the location-dependent relationship between richness and remote sensing imagery.

Geo-coordinates provide geographical priors that supplement geolocation context to the image data \cite{mai2022review, ge2022geoscience, tang2015improving, ayush2021geography, berg2014birdsnap, ardeshir2014gis, amlacher2009geo}. While a straightforward way to utilise geolocational features is to concatenate the original longitude and latitude coordinates into the model, this approach has shown to yield almost no gain in performance \cite{tang2015improving}. To deal with this problem, a geo-feature extraction approach was proposed in \cite{tang2015improving} for CNN models, where the geo-coordinates were projected into a higher dimensional feature space with a geolocation encoder, whose outputs were then merged into those of a CNN-based image network. It was observed that, by leveraging geolocation context, a 7\% increase in accuracy was achieved for an image data set spanning over the continental United States \cite{tang2015improving}. This geo-encoded CNN model was later applied to global-scale vegetation canopy height mapping with satellite imagery \cite{lang2023high}, where the geolocations served as a prior. Other state-of-the-art geolocation encoders include Space2Vec \cite{mai2020multi}, Sphere2Vec \cite{mai2023sphere2vec}, PE-GNN \cite{klemmer2023positional}, and a more recent algorithm that is based on the spherical harmonic basis functions \cite{russwurm2024geographic}. Multi-scale sinusoidal functions are favoured in building these encoders (\emph{e.g.}, \cite{mai2020multi, russwurm2024geographic}), thanks to their merits of being bounded in value, infinitely extended in space, and possessing a multi-resolution scalability. Geolocation encoding is demonstrated to be effective in many large-scale geospatial problems, such as animal species categorisation \cite{lang2023high, mai2023sphere2vec}, water quality prediction \cite{banerjee2022machine}, event/activity recognition \cite{joshi2008inferring}, and remote sensing scene classification \cite{ayush2021geography, mai2023sphere2vec}. 

In this study, we aim to predict the spatiotemporal distribution of plant species richness in Australia from EO imagery with geolocation context being taken into account. Considering that the relationship between plant species richness and remote sensing imagery varies from one location to another due to differences in vegetation composition, we propose \textit{Spatioformer}, where a novel geolocation encoder is coupled with the transformer model in order to incorporate the geolocation context. The performance of Spatioformer in richness mapping is compared with a CNN model, a ViT model, and the FactoFormer model where the geolocational information is not encoded. Through quantitative analyses, we seek to address primarily the following important questions: (1) Does Spatioformer perform better than state-of-the-art algorithms in predicting plant species richness over large spatial scales? (2) What are the spatial patterns of plant species richness in Australia inferred from remote sensing evidence? (3) Where should future in-situ surveys be conducted as suggested by the mapping results?

The rest of the paper is organised as follows. Section \ref{sec:study_area_and_datasets} describes the study area and the datasets used for modelling, including the ground-truth samples of plant species richness and satellite imagery. Section \ref{sec:methods} introduces the methods with a focus on the proposed Spatioformer model. This model is developed with the aim to capture the location-dependent relationships between plant species richness and remote sensing imagery over large spatial scales. Section \ref{sec:experimental} describes the experimental settings for training and validation of the Spatioformer model. Section \ref{sec:results_and_discussions} presents the results of applying Spatioformer to plant richness mapping across Australia, and discusses the implications of these findings for biodiversity conservation and future research directions. Finally, Section \ref{sec:conclusions} concludes the paper.
\section{Study Area and Datasets}
\label{sec:study_area_and_datasets}

\subsection{Study Area}

\begin{figure}[tb!]
\centering
\includegraphics[width=8.5cm]{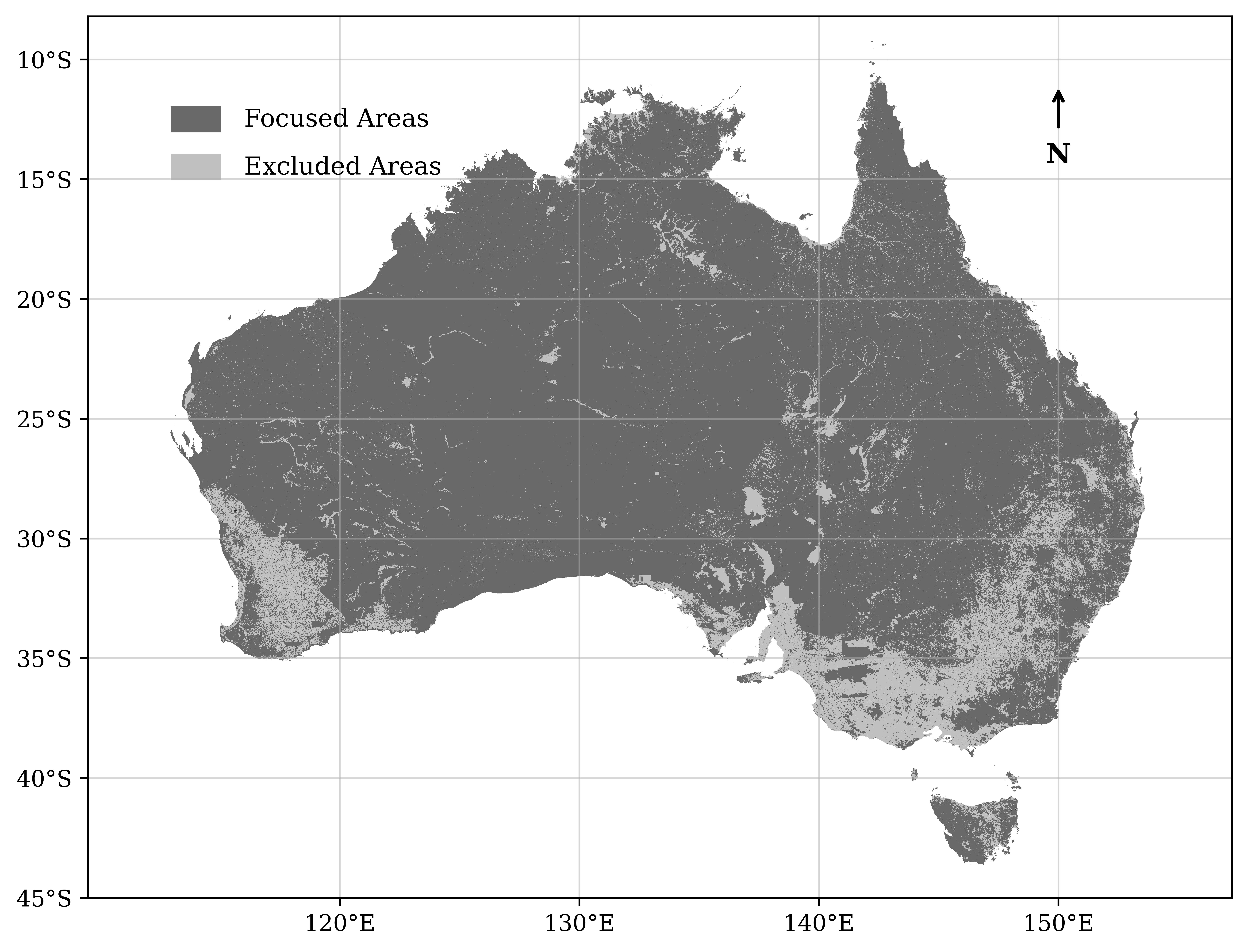}
\caption{Map of the study area. This work was focused on natural and near natural functioning terrestrial ecosystems within the Australian continent and nearby islands, as coloured in green in the figure, while heavily modified landscapes and water bodies coloured in grey were excluded from our analysis.\label{fig:aoi}}
\end{figure}

This study was focused on natural and near natural functioning terrestrial ecosystems within the Australian continent and nearby islands, as shown in Fig. \ref{fig:aoi}. The ecosystems considered include natural lands (\emph{e.g.}, conservation reserves, managed resource protected areas, and lands of minimal use), and lands utilised for production purposes from near natural environments (\emph{e.g.}, native grasslands and forests for production purposes) (Fig. \ref{fig:aoi}). We excluded heavily modified landscapes from our analysis, including agricultural lands (\emph{e.g.}, croplands, horticultural lands, and plantation forests), urban regions (\emph{e.g.}, industrial and residential lands), and water bodies (\emph{e.g.}, lakes, rivers, and reservoirs) (Fig. \ref{fig:aoi}). Those areas were excluded because vegetation surveys, which primarily aim to gather information for the preservation of native plant species, have been rarely conducted in heavily modified landscapes. The focused and excluded areas were identified with the Catchment Scale Land Use of Australia dataset in 50 m spatial resolution (updated in December 2018) \cite{abares2021catchment}.

\subsection{Plant Species Richness Samples}

\begin{figure}[tb!]
\centering
\includegraphics[width=8.87cm]{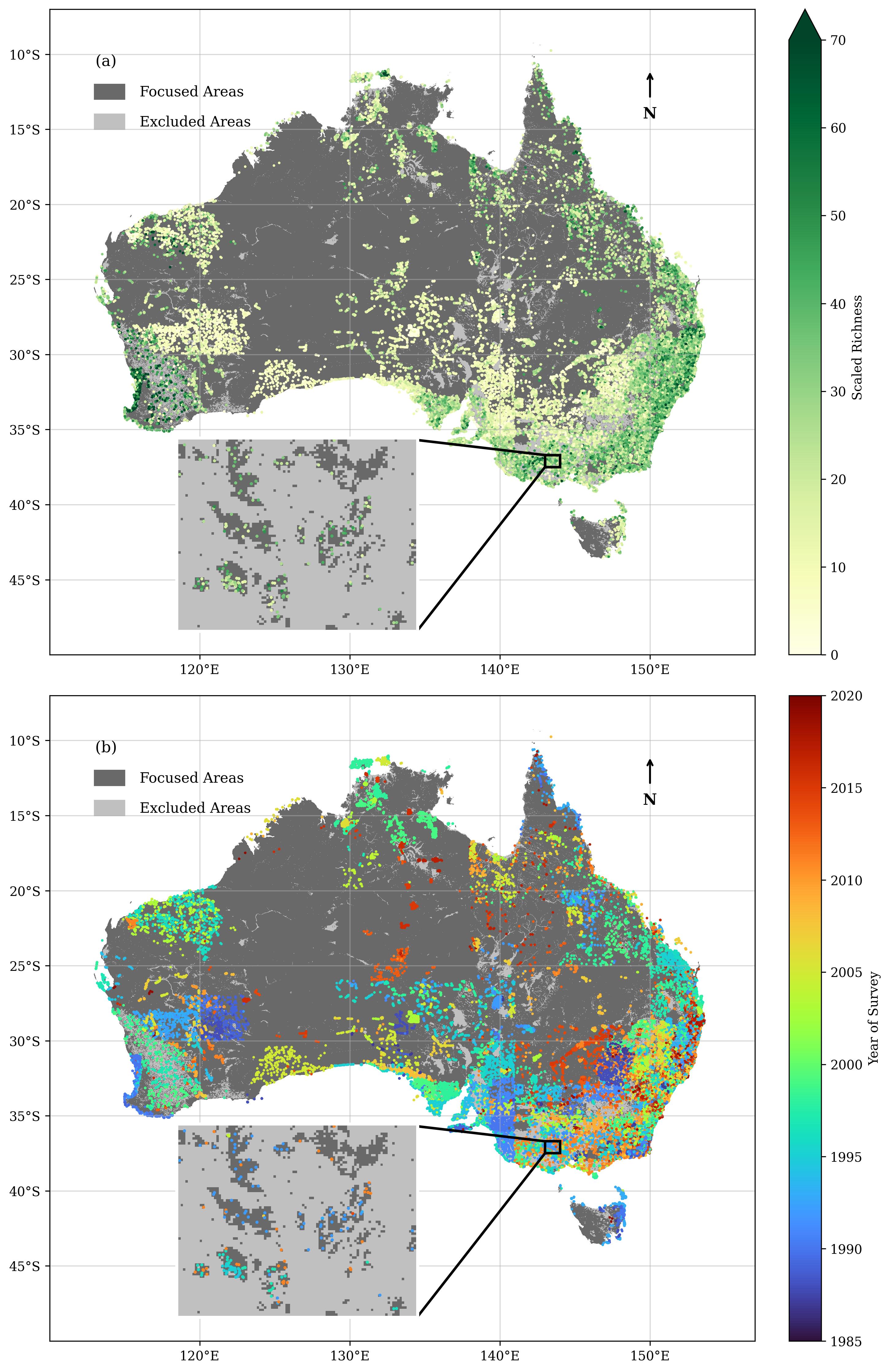}
\caption{Locations of ground survey samples, coloured by (a) species richness values (in unit of number of species per 400 m\textsuperscript{2}), and (b) years of survey. The insets provide zoomed-in views of a region in southeast Australia. A total of 68,170 samples from the Harmonised Australian Vegetation Plot (HAVPlot) dataset \cite{mokany2022harmonised, mokany2022patterns} were used for modelling in this study. These samples were collected via various field campaigns as a perseverant effort spanning the years from 1986 to 2020 (please refer to the Acknowledgements section for details on the custodians of these samples).\label{fig:location}}
\end{figure}

The ground-truth samples of plant species richness were obtained from the Harmonised Australian Vegetation Plot (HAVPlot) dataset \cite{mokany2022harmonised, mokany2022patterns}, where each sample represented a plot area of 400 m\textsuperscript{2} (20 m {\texttimes} 20 m). Out of the 219,552 samples in the HAVPlot dataset, 68,170 samples were selected for modelling in this study. These samples were collected via various field campaigns as a perseverant effort spanning the years from 1986 to 2020 (please refer to the Acknowledgements section for details on the custodians of these samples). As details in \cite{mokany2022patterns}, efforts have been made to harmonise the samples from different field campaigns, with the aim to minimise the temporal and spatial variations among samples arising from discrepancies in experimental design. The rest samples in the HAVPlot dataset were removed from our analysis because they consisted of less than 70\% native Australian species, or were collected before the year of 1986 and unable to be matched with a Landsat observation (as described later in Subsection \ref{ssec:satellite_imagery}). Fig. \ref{fig:location} shows the spatial locations of the selected species richness samples, coloured by the richness values (in unit of number of species per 400 m\textsuperscript{2}) (Fig. \ref{fig:location}a), and by the years of survey (Fig. \ref{fig:location}b). These samples cover diverse landscapes across Australia. Scattered pockets of natural or near-natural areas among heavily modified lands are also represented by samples (See the insets of Fig. \ref{fig:location}). Regions of easier human access, such as the southeast and southwest coastal areas, generally show a higher sample representativeness than the less habitable inland of arid or semi-arid climate. In particular, the arid/semi-arid western interior has low or no coverage of plots.

\subsection{Satellite Imagery}\label{ssec:satellite_imagery}

The satellite data were queried within CSIRO’s implementation of the Open Data Cube \cite{lewis2017australian} known as the Earth Analytics Science and Innovation (EASI) platform. Landsat images were extracted from the Geoscience Australia Landsat Geometric Median (Geomedian) and Median Absolute Deviation (MAE) Collection 3 product (version 3.10) \cite{norman2021dea}. The Geomedian Landsat-5 and 8 images with a spatial resolution of 30 m {\texttimes} 30 m were downloaded for years from 1986 to 2011 and from 2013 to 2023, respectively. Each pixel in the Geomedian imagery represented the statistical median of all observations for that pixel from the given calendar year \cite{roberts2017high,roberts2018open}. The geomedian observations provided a summarised  representation of annual habitat conditions for Australia's landscapes, considering that vegetation in the vast arid and semi-arid regions revive after each rainfall and exhibit no definite phenological cycles. These Geomedian images were analysis-ready surface reflectance data, with a series of pre-processing steps having been applied, including atmospheric correction, terrain correction, and Bidirectional Reflectance Distribution Function (BRDF) adjustment. The six spectral bands shared by Landsat-5 and 8 were selected for analysis, namely Blue, Green, Red, Near-Infrared, and Short-Wave-Infrared 1 and 2. For each richness sample, we matched it with a Geomedian Landsat image from the same year in which the sample was collected. The image consisted of 9 {\texttimes} 9 pixels with the centre pixel aligned to the location of the survey plot. We supplied spatially adjacent pixels neighbouring the plot for modelling, along with the centre pixel itself, because plants often live as a community where each plant interacts with its spatial neighbours in a close relationship. For example, the variation in spectral patterns among neighbouring pixels, known as the spectral diversity or optical diversity \cite{ustin2010remote}, is related to plant species diversity via the spectral variation hypothesis \cite{palmer2002quantitative}. The spatially neighbouring pixels thus may provide supplementary information to assist in the richness prediction.
\section{Methods}
\label{sec:methods}

\subsection{Overview}

To encode geospatial information into remote sensing imagery for large-scale plant species richness mapping, we propose the Spatioformer model which integrates the transformer with a geolocation encoder. In the following, we first introduce our approach to geolocation encoding in Subsection \ref{ssec:geolocational}, followed by detailing the Spatioformer structure in Subsection \ref{ssec:spatioformer}. Theoretical analysis of Spatioformer is given in Subsection \ref{ssec:training}. Finally, plant species richness mapping with Spatioformer is described in Subsection \ref{ssec:plant}.

\subsection{Geolocation Encoder}
\label{ssec:geolocational}

We choose multi-scale sinusoidal functions, \emph{i.e.}, sine and cosine with different frequencies, for geolocation encoding. This choice is supported by previous studies (\emph{e.g.}, \cite{mai2020multi, russwurm2024geographic}) that demonstrated the advantages of employing multi-scale sinusoidal functions for geolocation encoding, such as their bounded values, infinite spatial extension, and multi-resolution scalability. For a given geolocation $\left (x_i,y_i\right )$ with $x_i$ and $y_i$ being the longitude and latitude coordinates, its geolocation token is encoded as:
\begin{equation}
\bm{g}^{\left(x_i,y_i\right)}=[g_1^{\left(x_i,y_i\right)}, g_2^{\left(x_i,y_i\right)}, \cdots, g_j^{\left(x_i,y_i\right)}, \cdots,  g_d^{\left(x_i,y_i\right)}]^{\rm{T}},
\end{equation}
where $d$ is the dimension of geolocation token and ${\rm{T}}$ denotes transpose. The $j$th element in $\bm{g}^{\left(x_i,y_i\right)}$, $ g_j^{\left(x_i,y_i\right)}$, is calculated as follows:
\begin{equation}\label{eq:geolocational}
g_j^{\left(x_i,y_i\right)}=
\begin{cases}
\sin\left (\frac{x_i}{w_j} \right)+\sin\left (\frac{y_i}{v_j}\right ) \quad {\rm{if}} \; j \; {\rm{is}} \; {\rm{even}}, \\
\cos\left (\frac{x_i}{w_j} \right)+\cos\left (\frac{y_i}{v_j}\right ) \quad {\rm{if}} \; j \; {\rm{is}} \; {\rm{odd}},\\
\end{cases}
\end{equation}
where $w_j=a\cdot c^{\frac{j}{d}}$ and $v_j=a\cdot c^{\frac{d-j}{d}}$, with $a$ and $c$ being pre-set constants, determine the spatial frequencies of encoding layers. The constants $a$ and $c$ need to be set such that each pixel within the mapping area is provided with a sufficiently distinctive geolocation token by the encoding layers. 

\begin{figure}[tb!]
\centering
\includegraphics[width=8.87cm]{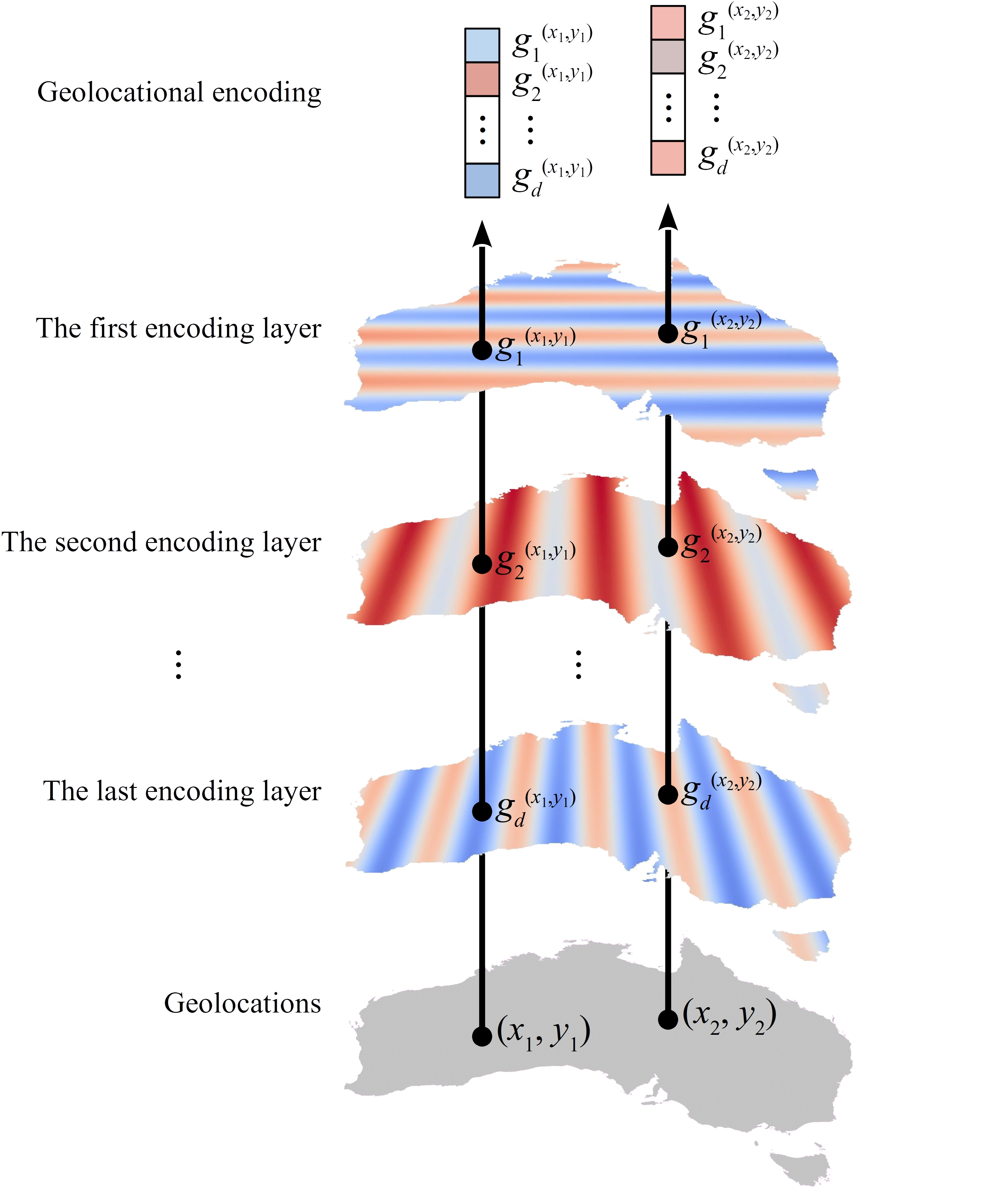}
\caption{A graphic illustration of geolocation encoding for two example locations $(x_1, y_1)$ and $(x_2, y_2)$. The geolocation encoding vectors for these two locations were constructed with values referenced from the corresponding positions on the encoding layers.\label{fig:encoding}}
\end{figure}

A graphic illustration of the proposed geolocation encoder is shown in Fig. \ref{fig:encoding} for two example locations. The geolocation encoding vectors for these two locations are constructed with values referenced from the corresponding positions in a total of $d$ encoding layers. This geolocation encoder could provide geo-gradient information not only in the cardinal directions but also in various diagonal directions (as shown in Fig. \ref{fig:geoencoding} in Subsection \ref{ssec:geolocational}). The $d$, $a$, and $c$ values configured in this study for plant species richness mapping are described in Subsection \ref{ssec:setup}.

\subsection{The Spatioformer Structure}
\label{ssec:spatioformer}

\begin{figure*}[tb!]
\begin{minipage}{.7\textwidth}
\centering
\includegraphics[width=11.5cm]{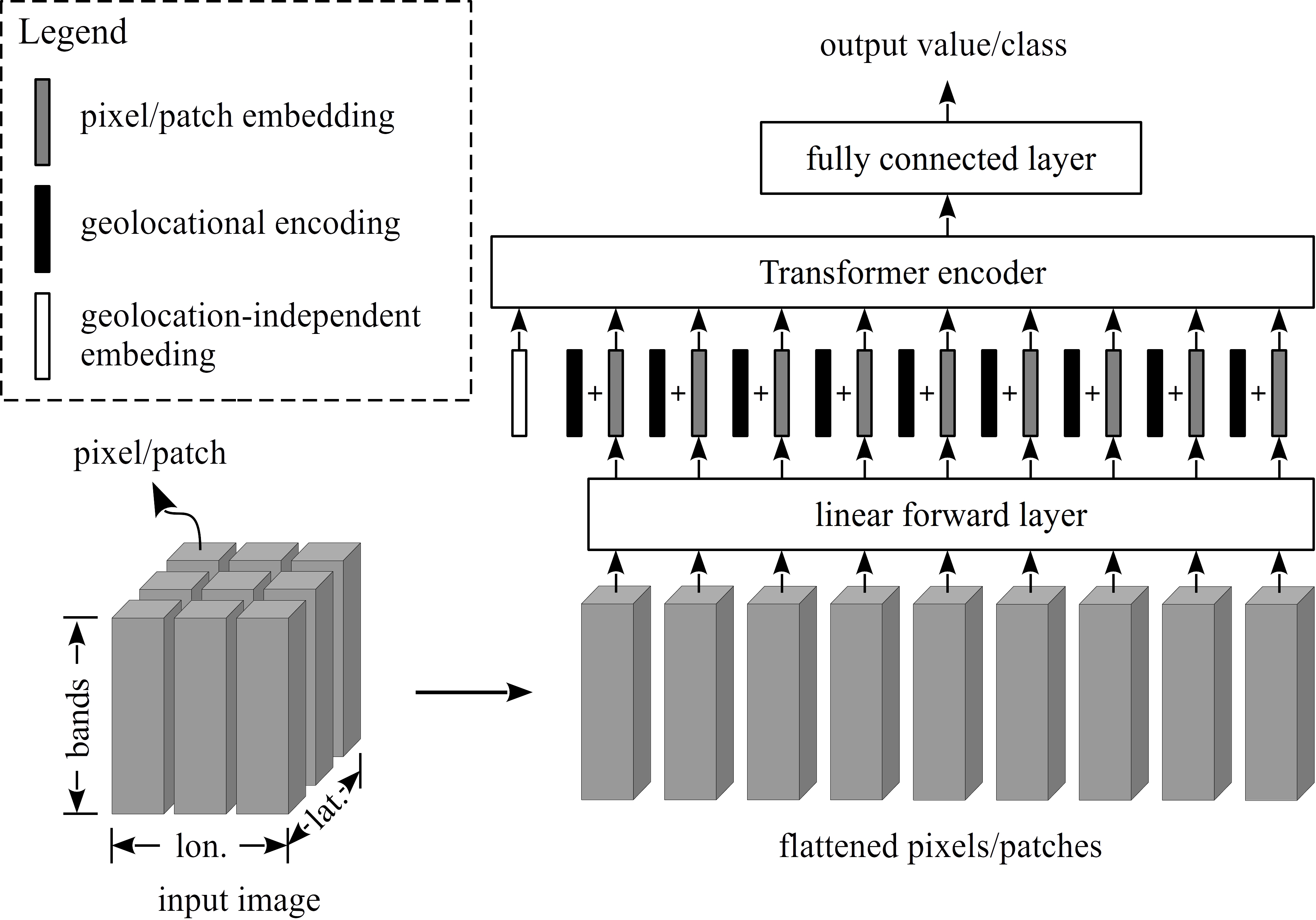}
\end{minipage}
\begin{minipage}{.3\textwidth}
\vspace{-3.7cm}
\caption{Graphic illustration of the Spatioformer structure. An image is first spatially divided into separate pixels or image patches, and then flattened, before being fed into a linear forward layer which projects the pixels/patches into the embedding space. For each pixel/patch, its embedding is added by its geolocation token. The geolocation-encoded embeddings are then fed into the transformer encoder, together with a geolocation-independent token to account for geolocation-independent components in the input-output relationship. A fully connected layer is set as the output layer to produce the predicted value or class.\label{fig:spatioformer}}
\end{minipage}
\end{figure*}

The Spatioformer structure is shown in Fig. \ref{fig:spatioformer}. The input of the model is an image in two spatial dimensions (\emph{i.e.}, longitude and latitude) and the spectral dimension. The image is first spatially divided into separate pixels or image patches, and then flattened, before being fed into a linear forward layer which projects the pixels/patches into the embedding space. For each pixel/patch, its embedding $\bm{e}_i$ is added by its geolocation token $\bm{g}^{\left(x_i,y_i\right)}$ which is calculated from its geo-coordinates ${\left(x_i,y_i\right)}$ (as described in Section \ref{ssec:geolocational}):
\begin{equation}
\label{eq:add}
\bm{e}'_i=\bm{e}_i+\lambda \bm{g}^{\left(x_i,y_i\right)},
\end{equation}
where $\lambda$ is a learnable parameter balancing the relative contribution between pixel values and geolocational information, given that these two different types of data may differ in their magnitudes. The geolocation-encoded embeddings $\bm{e}'_i$ are then fed into the transformer encoder, together with a learnable geolocation-independent token at the pixel/patch level to account for geolocation-independent components in the input-output relationship. A fully connected layer is set as the output layer to produce the predicted value or class, depending on whether the problem to be solved is a regression or classification problem. The configuration of Spatioformer for plant species richness mapping is described in Subsection \ref{ssec:setup}. The geolocation encoder in Spatioformer integrates geolocation context into the images, enabling the model to leverage this information in modelling plant species richness distribution in this study.

\subsection{Theoretical Analysis}
\label{ssec:training}

At the core of standard transformer models lies the self-attention module. For a single-head self-attention module \cite{ke2021rethinking}, with an input length of $n$ tokens, its output for the $i$th input token $\bm{x}_i$ can be expressed as:
\begin{equation}
\begin{split}
\label{eq:self-attention}
\bm{z}_i&=\sum_{j=1}^{n}\frac{{\rm{exp}}(\alpha_{ij})}{\sum_{j'=1}^{n}{\rm{exp}}(\alpha_{ij'})}(\bm{x}_j\bm{W}^V),\\ \alpha_{ij}&=\frac{1}{\sqrt{d}}(\bm{x}_i\bm{W}^Q)(\bm{x}_j\bm{W}^K)^{\rm{T}},
\end{split}
\end{equation}
where $\bm{W}^Q$, $\bm{W}^K$, and $\bm{W}^V$ are trainable weights for the Query, Key, and Value matrices, respectively.

In Spatioformer, the pixel embeddings are added together with their respective geolocation encoding tokens, before being fed into the self-attention module. Compared with concatenating longitude and latitude coordinates directly to pixel embeddings, the geolocation encoding is able to project geo-coordinates into a higher dimension (\emph{i.e.}, the pixel embedding space) as a better feature descriptor. As a result, the calculation of $\alpha_{ij}$ is different from that shown in Eq. (\ref{eq:self-attention}):
\begin{equation}
\label{eq:alpha}
\alpha_{ij}=\frac{1}{\sqrt{d}}[(\bm{x}_i+\lambda\bm{g}_i)\bm{W}^Q][(\bm{x}_j+\lambda\bm{g}_j)\bm{W}^K]^{\rm{T}},
\end{equation}
By expanding Eq. (\ref{eq:alpha}), we get:
\begin{equation}
\label{eq:expansion}
\begin{split}
\alpha_{ij}=&\underbrace{\frac{1}{\sqrt{d}}(\bm{x}_i\bm{W}^Q)(\bm{x}_j\bm{W}^K)^{\rm{T}}}_{\text{(1)}}
+
\underbrace{\frac{1}{\sqrt{d}}(\bm{x}_i\bm{W}^Q)(\lambda\bm{g}_j\bm{W}^K)^{\rm{T}}}_{\text{(2)}}\\
&+
\underbrace{\frac{1}{\sqrt{d}}(\lambda\bm{g}_i\bm{W}^Q)(\bm{x}_j\bm{W}^K)^{\rm{T}}}_{\text{(3)}}
+
\underbrace{\frac{1}{\sqrt{d}}(\lambda\bm{g}_i\bm{W}^Q)(\lambda\bm{g}_j\bm{W}^K)^{\rm{T}}}_{\text{(4)}},
\end{split}
\end{equation}
The four terms in the right-hand side of Eq. (\ref{eq:expansion}) correspond to the pixel-to-pixel (Term 1), pixel-to-geolocation (Term 2), geolocation-to-pixel (Term 3), and geolocation-to-geolocation attention (Term 4), respectively. Therefore, training Spatioformer is equivalent to joint optimisation of pixel embedding and geolocation encoding by taking into account their respective self-correlations and the correlations between each other.

\subsection{Plant Species Richness Mapping}
\label{ssec:plant}

In this study, the Spatioformer model was trained with ground-truth richness samples and their corresponding Landsat images. With the trained model, plant species richness maps were compiled through inferring richness values from Landsat images. Mapping results were demonstrated for the years from 2015 to 2023. Compared with older years, maps for these recent years can better inform current and future conservation activities and policy making. We then aggregated these annual maps to compile the mean richness map across the nine years in order to study plant species richness with enhanced spatial patterns. We also calculated the standard deviation map of richness over the nine years to identify places where predicted plant species richness showed high temporal variations. The size of input images was ranged from 1 {\texttimes} 1 to 9 {\texttimes} 9 pixels in order to examine how mapping accuracy varied when different spatial scales of plant community were taken into account.

The uncertainty map for richness predictions was produced with the Monte Carlo Dropout approach \cite{gal2016dropout}. For each predicted richness value, its uncertainty was calculated as the coefficient of variation over multiple additional model predictions under a dropout rate of 0.5 that randomly drops 50\% of model parameters, as follows:
\begin{equation}
\varepsilon = \sqrt{\frac{\sum_{i=1}^{n}(y_i-\bar{y})^2}{n-1}} \mathbin{/} \bar{y}, \quad \quad \bar{y}=\frac{\sum_{i=1}^{n}y_i}{n},
\end{equation}
where $\varepsilon$ is the uncertainty metric, $y_i$ is the predicted species richness under the dropout mode, and $n$ is the number of additional model predictions. In this study, $n$ was set to 100. 
\section{Experimental Setup}
\label{sec:experimental}

\subsection{Dataset Partition}

In this study, the in-situ samples across all years were used to build a model under a training/validation/test data splitting scheme, followed by applying the model to make mapped predictions for each year. A random data splitting scheme for accuracy assessment in large-scale mapping was cautioned by \cite{karasiak2022spatial, kattenborn2022spatially, ploton2020spatial}, as randomly sampled hold-outs of validation and test data may spatially autocorrelated with the training samples, leading to an overestimation of accuracy. Data splitting schemes based on spatially blocked hold-outs were therefore recommended to mitigate this problem \cite{kattenborn2022spatially, ploton2020spatial}. 

\begin{figure}[tb!]
\centering
\includegraphics[width=8.8cm]{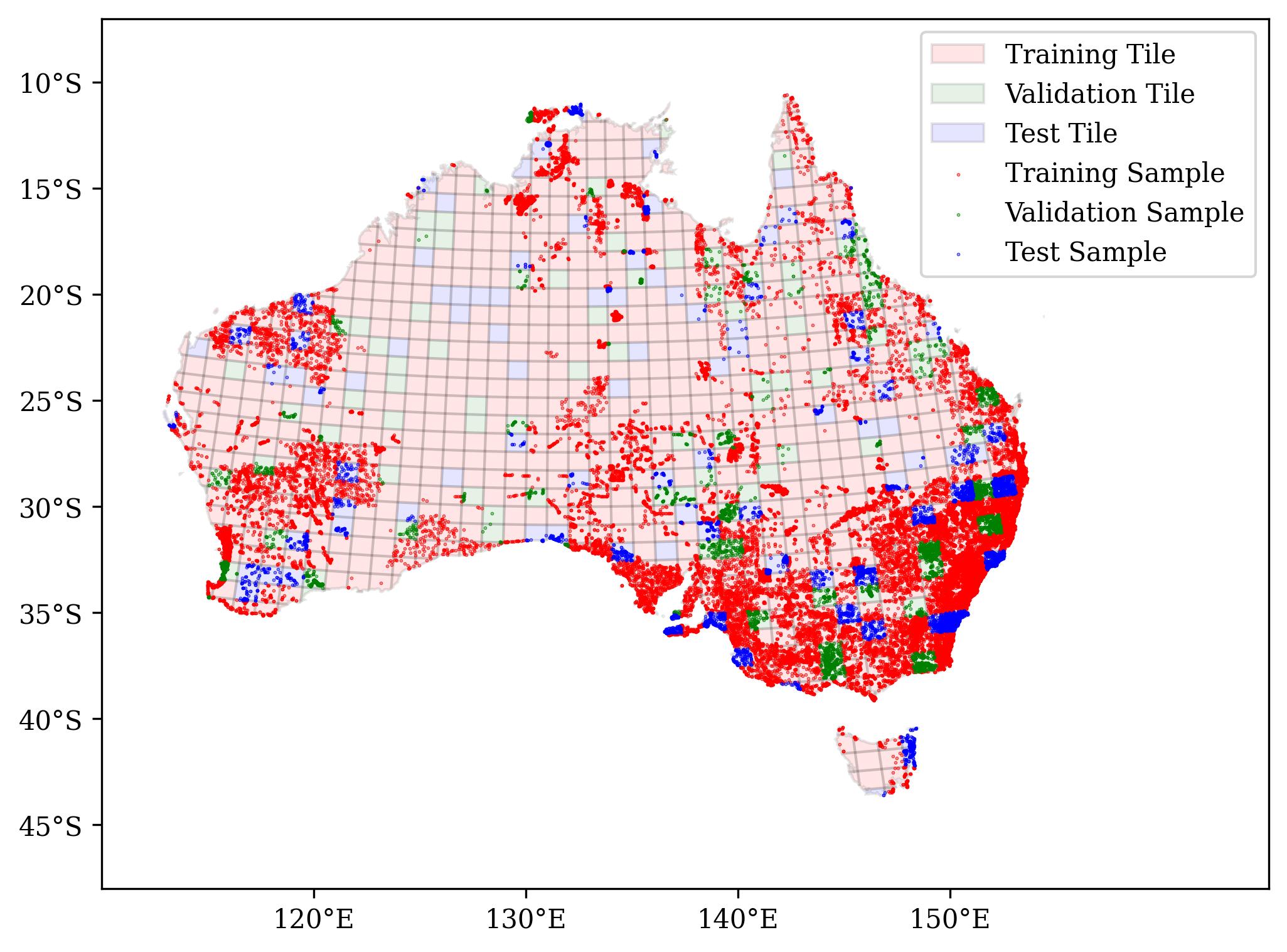}
\caption{Partition of ground samples
into training, validation, and test sets based on geographical tiles. The Australian territory was divided into 958 tiles of 100 km {\texttimes} 100 km, with 766 tiles (approx. 80\%), 96 tiles (approx. 10\%), and 96 tiles (approx. 10\%) being randomly selected as the training, validation, and test tiles. Samples located within the training, validation, and test tiles were assigned into the respective sets. \label{fig:split}}
\end{figure}

In our work, a block-based training/validation/test scheme was adopted. We first divided the Australian territory into 958 tiles of 100 km {\texttimes} 100 km based on a geographical zoning scheme provided by Geoscience Australia, as shown in Fig. \ref{fig:split}. Among them, we randomly selected 766 training tiles (approx. 80\%), 96 validation tiles (approx. 10\%), and 96 test tiles (approx. 10\%) (Fig. \ref{fig:split}). Ground-truth richness samples located within the training, validation, and test tiles were assigned into the respective sets (Fig. \ref{fig:split}). As a result, the training, validation, and test sets consisted of 55159, 6753, and 6258 samples, respectively. In addition to mitigating the spatial autocorrelation problem, having training, validation, and test sets spatially separate from each other could also help the Spatioformer avoid overfitting to local data anomalies, which would otherwise generate hotspot artifacts (\emph{i.e.,} extreme values at local scales due to overfitting) in the predicted richness maps.

\subsection{Spatioformer Model Setup and Hyperparameter Selection}
\label{ssec:setup}

The input images of 9 {\texttimes} 9 pixels, with each pixel being of 30 m {\texttimes} 30 m spatial resolution, were embedded into a 16-dimensional feature space using a linear feed-forward layer. In cases where input images have irregular edges, a patching strategy can be implemented by applying zero-padding along the edges to standardise the input dimensions. A geolocation token of 16 dimensions was added to the embedding of each pixel, where the embedding feature dimension of 16 was determined based on multiple test runs. The geolocation-encoded pixel embeddings were fed into three layers of eight-head self-attention module with each layer followed by a 64-dimensional feed-forward network. A fully connected network with a hidden layer of 1024 nodes was set as the output layer. The mean squared error between model-predicted and ground-truth richness values was employed as the loss function. The Adam optimizer was employed for network optimisation.

Model hyperparameters were determined with multiple test runs. The learning rate and weight decay were set to 1\texttimes 10\textsuperscript{-3} and 1\texttimes 10\textsuperscript{-4}, respectively. A cosine annealing scheme with warm start was scheduled for the learning rate. A dropout rate of 0.1 was applied. The constants $a$ and $c$ in geolocation encoder were set to 1 and 100, respectively. The initial value for the learnable parameter $\lambda$ balancing the relative contribution between pixel values and geolocational information was set to 1\texttimes 10\textsuperscript{4}.

\subsection{Benchmarking with State-of-the-Art Models}

The performance of Spatioformer was compared with a CNN model \cite{lecun1998gradient}, a ViT model \cite{vaswani2017attention}, and the FactoFormer model \cite{mohamed2023factoformer}. The CNN model consisted of three convolutional layers, with each layer followed by a ReLU activation layer and a batch normalisation layer. Each convolutional layer consisted of eight filters with a kernel size of 3 {\texttimes} 3. Following the convolutional layers were a flatten layer and a fully connected network. For the ViT model, each pixel was first embedded into a 16-dimensional feature space via a linear feed-forward layer, which was then fed into three layers of eight-head self-attention module. Each self-attention layer was followed by a feed-forward network of 64 dimensions. A flatten layer and a fully connected network were set as the output layers. For the FactoFormer model, each input image cube was first split into non-overlapping tokenized patches along spectral and spatial dimensions, followed by processing them with two transformers simultaneously. Then the outputs of each transformer were flatten, concatenated, and passed to a a fully connected network. 

These benchmark models were optimised on their performance in the same way as the optimisation of Spatioformer structure, \emph{i.e.}, grid search of the optimal model hyperparameters via multiple test runs. The same as the setting for Spatioformer, 9 {\texttimes} 9 image pixels over each richness sample were supplied to CNN, ViT, and FactoFormer as input. Seven evaluation metrics were calculated and compared for these models, including coefficient of correlation (\textit{r}), coefficient of determination (\textit{r}\textsuperscript{2}), Mean Absolute Error (MAE), Relative Absolute Error (RAE), Mean Squared Error (MSE), Relative Squared Error (RSE), and Root Mean Squared Error (RMSE) between model-predicted and ground-truth species richness (in unit of number of species per 400 m\textsuperscript{2}).

\section{Results and Discussions}
\label{sec:results_and_discussions}

\subsection{Geolocation Encoding}
\label{ssec:encoding}

\begin{figure*}[ht!]
\centering
\includegraphics[width=17cm]{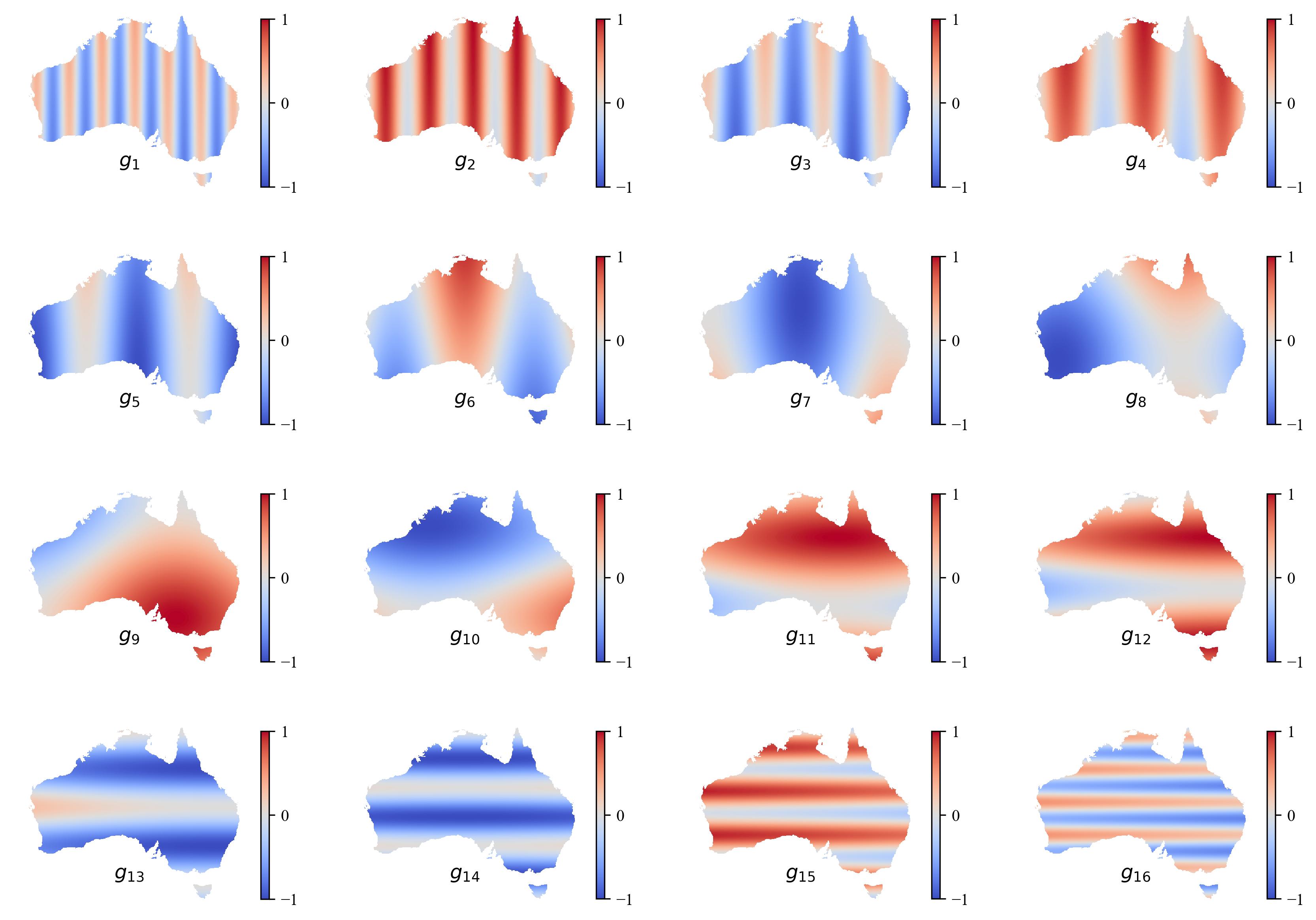}
\caption{Geolocation encoding layers for plant species richness mapping in our study. These layers were generated with the proposed geolocation encoder. The first a few layers (from \textit{g}\textsubscript{1} to \textit{g}\textsubscript{4}) highlight the longitudinal gradients, the last a few layers (from \textit{g}\textsubscript{13} to \textit{g}\textsubscript{16}) highlight the latitudinal gradients, and the middle layers (from \textit{g}\textsubscript{5} to \textit{g}\textsubscript{12}) show smoother spatial pattern in diagonal directions.\label{fig:geoencoding}}
\end{figure*}

The 16 geolocation encoding layers adopted in our study for plant species richness prediction are shown in Fig. \ref{fig:geoencoding}. By comparing the spatial patterns of these layers, it was observed that the first a few layers (from \textit{g}\textsubscript{1} to \textit{g}\textsubscript{4}) highlighted the longitudinal gradients of different spatial frequencies, while the last a few layers (from \textit{g}\textsubscript{13} to \textit{g}\textsubscript{16}) the latitudinal gradients. The middle layers (from \textit{g}\textsubscript{5} to \textit{g}\textsubscript{12}) showed smoother spatial patterns that presented diagonal gradients in various directions. The differed spatial patterns rendered by these layers were capable of providing adequate discriminating power to identify the encoding of one location from that of another. 

In richness prediction, each geo-location was encoded as a 16-dimensional token whose elements were referenced from the corresponding positions in the encoding layers shown in Fig. \ref{fig:geoencoding}. These tokens could project the geo-coordinates into a higher dimensional feature space,  given that utilising the original longitude and latitude coordinates as features was shown to yield almost no gain in performance \cite{tang2015improving}. As observed from the figure, the high-dimensional encoding could provide geo-gradient information not only in the cardinal directions but also in various diagonal directions. The encoding layers with information of varied spatial frequencies also made it possible to distinguish geo-information at different spatial scales. 

The geolocation tokens in richness prediction aimed at providing sufficient geolocation context such that the location dependency in the relationship between plant species richness and remote sensing imagery could be accounted for. Considering the continental scale of plant species richness mapping that this study aimed to handle, the geolocation context might help the model distinguish locations with dissimilar plant compositions, with the potential to increase the richness prediction accuracy across large spatial scales.

\subsection{Annual Richness Distributions}
\label{ssec:mapping_results}

\begin{figure*}[htb!]
\centering
\captionsetup{font=normalsize}
\includegraphics[width=18.1cm]{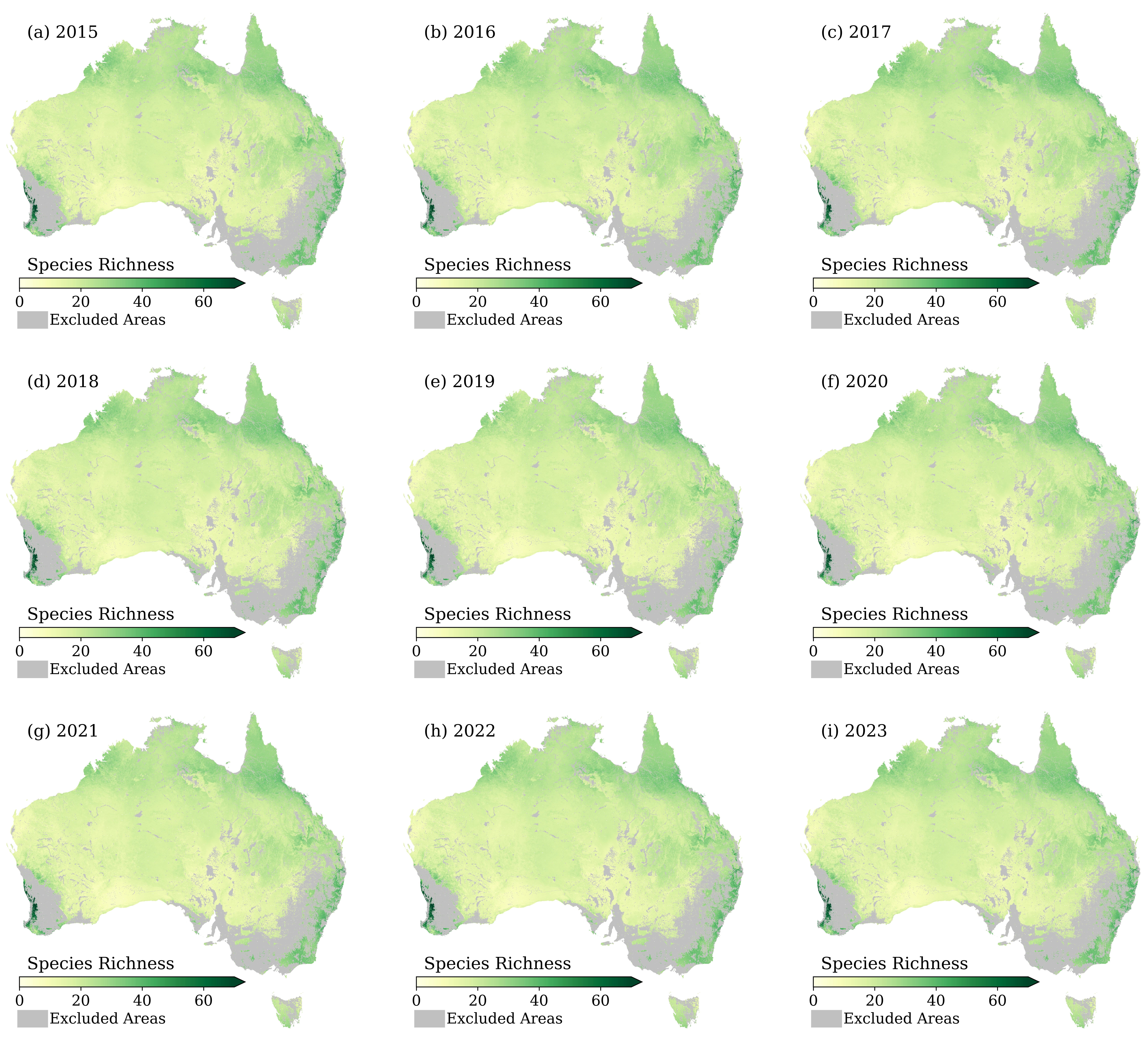}
\caption{Predicted annual maps of plant species richness distribution in Australia from 2015 to 2023 (a--i). These maps were compiled by applying the proposed Spatioformer model to Landsat observations from the respective years over the Australian territory. Richness values are in unit of number of species per 400 m\textsuperscript{2}.\label{fig:map}}
\end{figure*}

Fig. \ref{fig:map} shows the annual maps of plant species richness distribution in Australia from 2015 to 2023, which were compiled by applying the trained Spatioformer model to Landsat observations from the respective years. It was observed that the spatial patterns of Landsat-derived richness as displayed in these maps were consistent with the ground-sampled richness values displayed in Fig. \ref{fig:location}a. 

From the annual maps shown in Fig. \ref{fig:map}, it was seen that the Jarrah Forest bioregion near the southwestern coast of Western Australia showed the highest plant species richness within Australia, with values generally higher than 60 species per 400 m\textsuperscript{2}. The high species richness in this region could be evidenced by the high richness samples recorded in field surveys (Fig. \ref{fig:location}a). With a Mediterranean climate, this region harbours a biodiverse ecosystem with a distinctive composition of native vascular plants including the endemic Eucalyptus species \emph{Eucalyptus laeliae}, making it a region of high conservation value.

The eastern and southeastern coastal areas of Australia showed a medium-high richness of plant species with values ranging from 30 to 60 species per 400 m\textsuperscript{2} (Fig. \ref{fig:map}). These areas cover mostly the pristine ecosystems within the Great Dividing Range, where the climate varies from temperate to subtropical. Many vascular plant species of high conservation importance have been identified in these areas. For example, the Snowy Mountains region, located to the southwest of Canberra, is home to 212 species of vascular plants, of which 21 are endemic \cite{pickering2008vascular}. Another example is the Southern Tablelands region to the southwest of Sydney, where out of the approx. 1200 species of vascular plants, 30 have been listed as threatened due to a high degree of human interference and habitat alteration \cite{fallding2002planning}.

It was observed that the Australian Savanna in the north of continent (\emph{e.g.}, the northern parts of Western Australia, Northern Territory, and Queenland) showed a range of richness from 20 to 50 species per 400 m\textsuperscript{2} (Fig. \ref{fig:map}). The tropical/subtropical savanna climate in this region provides a copious amount of rainfall in the wet season, followed by a long dry season during the rest of the year. This unique climatic pattern results in a landscape that is characterised by broad grassland interspersed with small trees and shrubs. It was identified from Fig. \ref{fig:map} that the Kimberley Tropical Savanna (along the northwest coast of Kimberley), the Carpentaria Tropical Savanna (along the south coast of the Gulf of Carpentaria), and the Einasleigh Uplands Savanna (in mid-north Queensland) showed a relatively higher richness of plant species than other areas within the Australian Savanna.

The inland areas showed a low richness of plant species with the number of species per 400 m\textsuperscript{2} lower than 20 (Fig. \ref{fig:map}). This observation is in line with the range of richness values sampled in interior Australia (Fig. \ref{fig:location}a). The low richness could be mainly attributed to the arid/semi-arid climate that prevents the growth and sustenance of vascular plants, apart from the xeromorphic species. Although most coastal regions showed high to medium richness, the Nullarbor Plain (along the northwest coast of the Great Australian Bight) and the 
westernmost coastal region in Western Australia showed a very low richness of less than 10 species per 400 m\textsuperscript{2}. These two coastal regions are of limited plant diversity mainly because of the arid climate.

Compared with measuring richness via in-situ ground sampling, the EO-derived richness maps in Fig. \ref{fig:map} provided a more complete spatiotemporal coverage. This spatially continuous and up-to-date knowledge of richness distribution could serve as a valuable asset to inform effective conservation strategies and activities. It is worth noting that heavily modified landscapes (\emph{e.g.}, agricultural and urban regions) were masked out in the annual richness maps in Fig. \ref{fig:map}. It was because the in-situ data set, on which our model was trained, consisted mainly of samples from natural and near-natural landscapes; thus our model was not representative of heavily modified landscapes. Water bodies were also excluded from these richness maps as this work focused on terrestrial plants only. 

\subsection{Model Performance}

Table \ref{tab:accuracy} compares the performance of CNN \cite{lecun1998gradient}, ViT \cite{vaswani2017attention}, FactoFormer \cite{mohamed2023factoformer}, and the proposed Spatioformer model in predicting ground-truth species richness from the test dataset. It was seen from the figure that the CNN model achieved 0.52, 0.27, 10.41, 0.37, 188.69, 0.18, and 13.74 on \textit{r}, \textit{r}\textsuperscript{2}, RAE, MSE, RSE, and RMSE, respectively, between model-predicted and ground-truth species richness. The ViT model achieved 0.55, 0.29, 10.37, 0.37, 182.81, 0.17, and 13.52, and the FactoFormer model achieved 0.60, 0.35, 9.81, 0.35, 168.03, 0.16, and 12.96  on these metrics. The Spatioformer model compared favourably to the aforementioned benchmark models, showing 0.77, 0.59, 7.83, 0.29, 105.85, 0.11, and 10.29 on \textit{r}, \textit{r}\textsuperscript{2}, RAE, MSE, RSE, and RMSE, respectively. As the location-dependent relationship between satellite observations and on-ground richness values was accounted for in our modelling with Spatioformer, the results shown in Table \ref{tab:accuracy} suggested that the geolocation context could help improve prediction accuracy in plant species richness prediction. 

\begin{table*}[tb!]
\centering
\caption{Comparison between the proposed Spatioformer and benchmark models in predicting plant species richness. Evaluation metrics include coefficient of correlation (\textit{r}), coefficient of determination (\textit{r}\textsuperscript{2}), Mean Absolute Error (MAE), Relative Absolute Error (RAE), Mean Squared Error (MSE), Relative Squared Error (RSE), and Root Mean Squared Error (RMSE) between model-predicted and ground-truth plant species richness.\label{tab:accuracy}}
\renewcommand{\arraystretch}{1.3} 
\begin{tabular}{p{2.5cm}p{2.5cm}p{2.5cm}p{2.5cm}p{2.5cm}}
\hline
Metric      & CNN \cite{lecun1998gradient}   & ViT \cite{vaswani2017attention}  & FactoFormer \cite{mohamed2023factoformer}  & Spatioformer (Ours)    \\ \hline
\textit{r}  & 0.52   & 0.55  &0.60 & 0.77   \\
\textit{r}\textsuperscript{2} & 0.27   & 0.29 &0.35   & 0.59   \\
MAE         & 10.41  & 10.37  &9.81 & 7.83   \\
RAE         & 0.37   & 0.37   &0.35 & 0.29   \\
MSE         & 188.69 & 182.81 &168.03 & 105.85 \\
RSE         & 0.18   & 0.17   &0.16 & 0.11   \\
RMSE        & 13.74  & 13.52  &12.96 & 10.29  \\ \hline
\end{tabular}
\end{table*}

\begin{figure*}[htb!]
\centering
\captionsetup{font=normalsize}
\includegraphics[width=18.1cm]{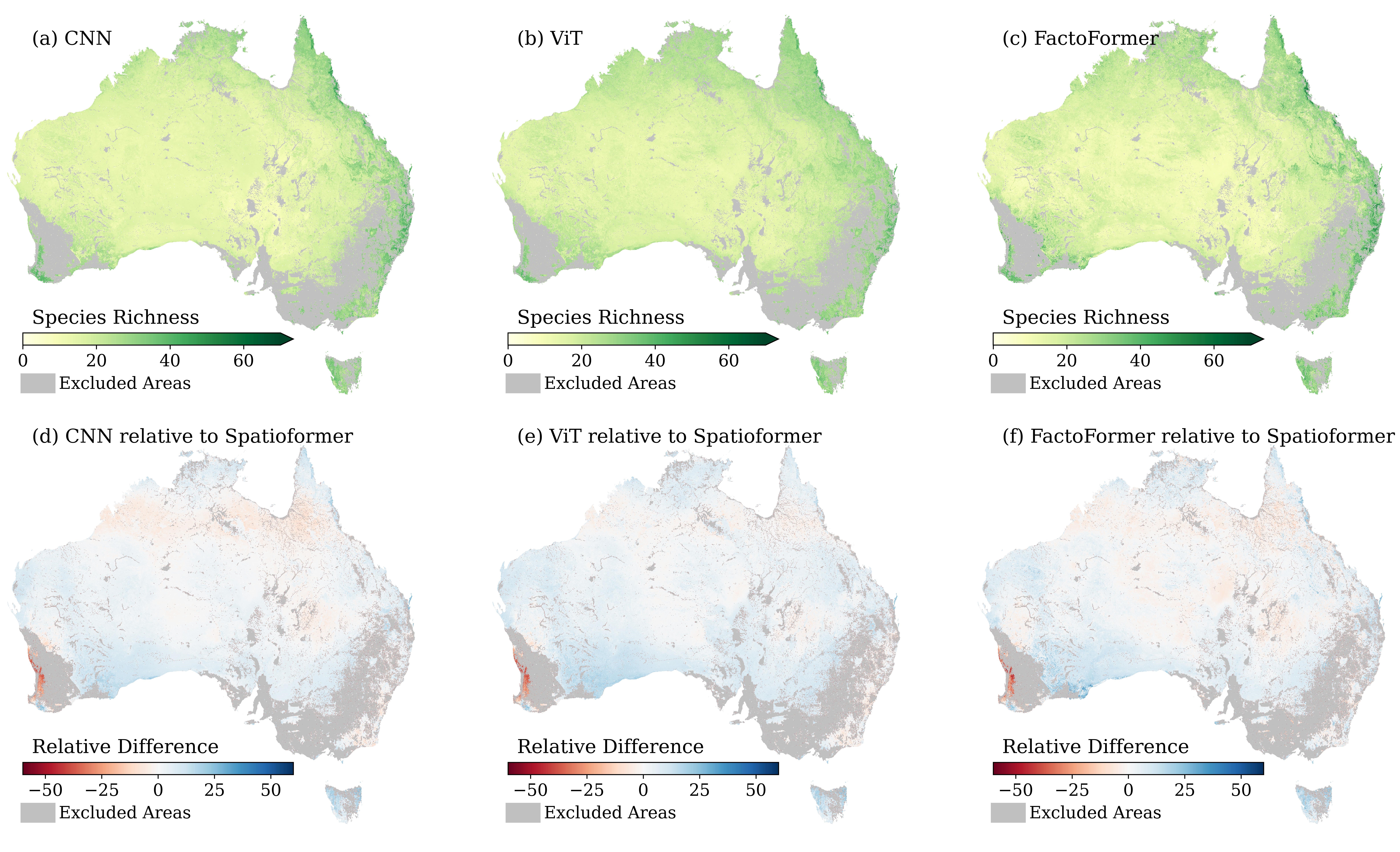}
\caption{Predicted plant species richness distribution in Australia with (a) Convolutional Neural Network (CNN) \cite{lecun1998gradient}, (b) Vision Transformer (ViT) \cite{vaswani2017attention}, and (c) FactoFormer \cite{mohamed2023factoformer}, for the year 2020, and (d--f) their relative difference to the result obtained with the proposed Spatioformer for the same year (Fig. \ref{fig:map}f). Richness values are in unit of number of species per 400 m\textsuperscript{2}.\label{fig:benchmark}}
\end{figure*}

\begin{figure}[htb!]
\centering
\captionsetup{font=normalsize}
\includegraphics[width=8.8cm]{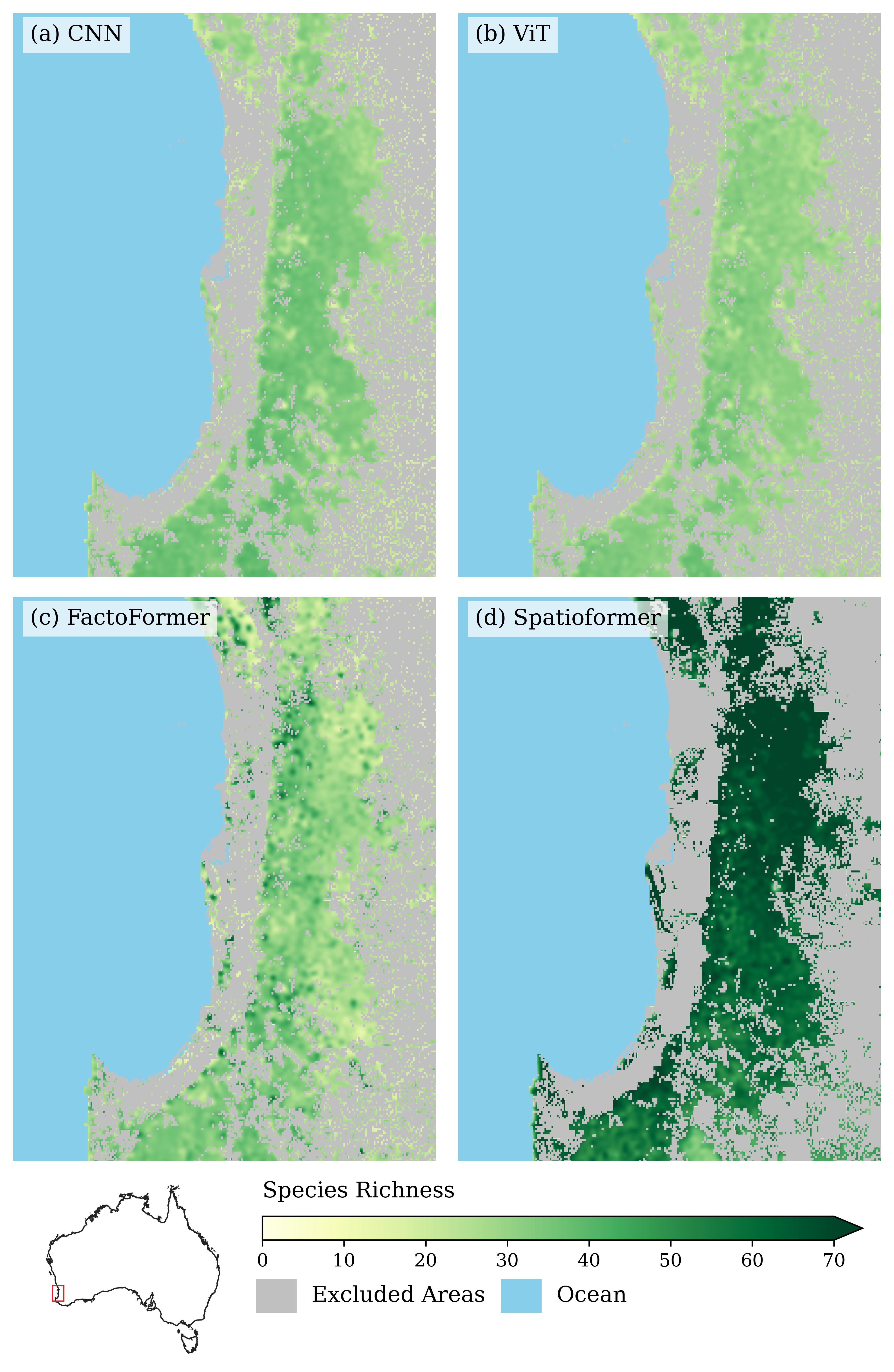}
\caption{Predicted plant species richness distribution in the Jarrah Forest bioregion near the southwestern coast of Western Australia with (a) Convolutional Neural Network (CNN) \cite{lecun1998gradient}, (b) Vision Transformer (ViT) \cite{vaswani2017attention}, (c) FactoFormer \cite{mohamed2023factoformer}, and (d) the proposed Spatioformer for the year 2020. Richness values are in unit of number of species per 400 m\textsuperscript{2}.\label{fig:zoomin}}
\end{figure}

Fig. \ref{fig:benchmark} shows the predicted distribution of plant species richness across Australia in the year 2020, generated by the three benchmark models: CNN \cite{lecun1998gradient} (Fig. \ref{fig:benchmark}a), ViT \cite{vaswani2017attention} (Fig. \ref{fig:benchmark}b), and FactoFormer \cite{mohamed2023factoformer} (Fig. \ref{fig:benchmark}c). The relative difference of these maps to the map produced by the proposed Spatioformer model (Fig. \ref{fig:map}f) reveals notable discrepancies in certain regions, as shown in Fig. \ref{fig:benchmark}d--f. For instance, the Jarrah Forest bioregion near the southwestern coast of Western Australia is a biodiverse region with richness generally higher than 60 species per 400 m\textsuperscript{2}, as evidenced by the ground-truth samples (Fig. \ref{fig:location}a). While the high richness in this region was successfully mapped with the proposed Spatioformer model (Fig. \ref{fig:map}f), the three benchmark models predicted this region to have 30 $\sim$ 40 species per 400 m\textsuperscript{2} (Fig. \ref{fig:benchmark}a--c), roughly 20 $\sim$ 40 lower relative to the Spatioformer predictions (Fig. \ref{fig:benchmark}d--f). The difference in richness prediction results between Spatioformer and the benchmark models is shown in more detail in Fig. \ref{fig:zoomin} with enlarged maps of the predicted plant species richness distribution in the Jarrah Forest bioregion. As observed from Fig. \ref{fig:zoomin}, the proposed Spatioformer model predicted the richness to be generally higher than 60 species per 400 m\textsuperscript{2} (Fig. \ref{fig:zoomin}d), matching better with the ground-truth evidence (Fig. \ref{fig:location}a) than the benchmark models which estimated roughly 30 to 40 species per 400 m\textsuperscript{2} (Fig. \ref{fig:zoomin}a--c). Another example is the Nullarbor Plain along the northwest coast of the Great Australian Bight. The extremely dry climate in this region has resulted in poor plant diversity of less than 10 species per 400 m\textsuperscript{2}, as supported by field surveys shown in Fig. \ref{fig:location}a. While the Spatioformer model accurately predicted the very low richness values observed in the Nullarbor Plain (Fig. \ref{fig:map}f), the predicted richness values by three benchmark models (Fig. \ref{fig:benchmark}a--c) were not as low as those indicated by the ground-truth samples, showing about 10 $\sim$ 20 higher relative to the Spatioformer map (Fig. \ref{fig:benchmark}d--f). These results suggested that, the proposed Spatioformer model is capable of effectively adapting to local richness characteristics, thanks to the incorporation of geolocation context in its modelling. 

As regions with a similar level of richness may present quite different spectral features in remote sensing imagery, predicting plant species richness over large spatial scales is a challenging task. Previous studies suggested that the relationship between plant species richness and remote sensing observations is location-specific, primarily due to difference in the assemblages of plant species among regions that results from their unique biogeographic histories, climate conditions, and geographical features \cite{rosenzweig1995species}. For example, statistical analysis showed that the Southern Tablelands and Snowy Mountains regions differ in their relationships between spectral data and plant species richness \cite{guo2023plant}. Therefore, geolocation context might be useful for mapping richness across regions. The location-dependent modelling with Spatioformer encoded such geolocational information into the input spectral data, which could help accommodate the location dependency in mapping richness over large spatial scales. The results shown in Table \ref{tab:accuracy}, the visual comparison between Fig. \ref{fig:benchmark} and Fig. \ref{fig:map}f, and the enlarged comparison maps shown in Fig. \ref{fig:zoomin}, indicated that it is beneficial to utilise geolocation context in large-scale richness mapping, with a higher overall accuracy being observed.

\subsection{Cross-Year Average and Standard Deviation of Richness}

\begin{figure}[tb!]
\centering
\includegraphics[width=8.5cm]{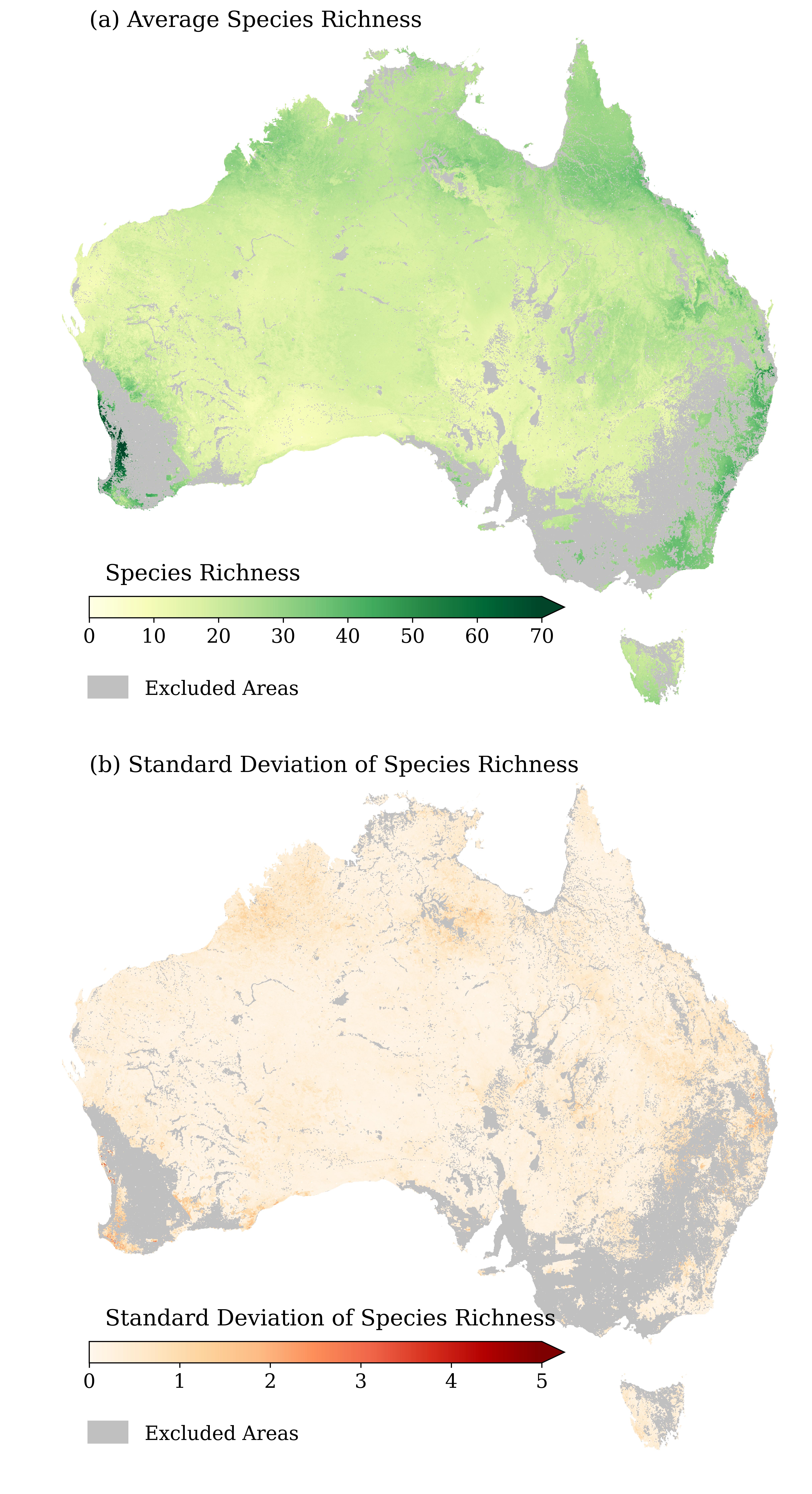}
\caption{(a) The average plant species richness map in Australia from 2015 to 2023, and (b) the standard deviation map of richness across these years derived from annual richness maps compiled with the proposed Spatioformer model.\label{fig:avg}}
\end{figure}

The maps in Fig. \ref{fig:avg} show the cross-year average and standard deviation of plant species richness in Australia, with the average map (Fig. \ref{fig:avg}a) showing the spatially enhanced long-term distribution of richness in Australia, and the standard deviation map (Fig. \ref{fig:avg}b) indicating the cross-year stability of richness across the country.  

It was observed from Fig. \ref{fig:avg}b that, the richness values were temporally stable with a standard deviation lower than one for most parts of the country, while higher cross-year variations were identified for several regions. Part of the southwestern, eastern, and southeastern coastal regions, as well as some areas in the northern savanna, showed a cross-year standard deviation in richness higher than one (Fig. \ref{fig:avg}b). Due to the relatively high biomass, these regions are susceptible to natural bushfires, such as the 2019--2020 Black Summer bushfires. In addition to natural bushfires, controlled prescribed burning has been practised in these regions aiming to reduce the risk of natural bushfires by intentionally burning excess flammable materials (\emph{e.g.}, dead woods). Floods, droughts, and human interference are also among the reasons that may contribute to cross-year variations in the richness levels.

With abrupt temporal variations and random noise being reduced, the average richness map in Fig. \ref{fig:avg}a provided a spatially enhanced reference for applications where the long-term distribution of richness needs to be taken into consideration. The standard deviation of richness in Fig. \ref{fig:avg}b provided information on cross-year richness stability for different parts of the country, which could serve as a reference in conservation planning and practice.

\subsection{Uncertainty in Richness Prediction}

\begin{figure*}[tb!]
\begin{minipage}{.7\textwidth}
\centering
\includegraphics[width=11cm]{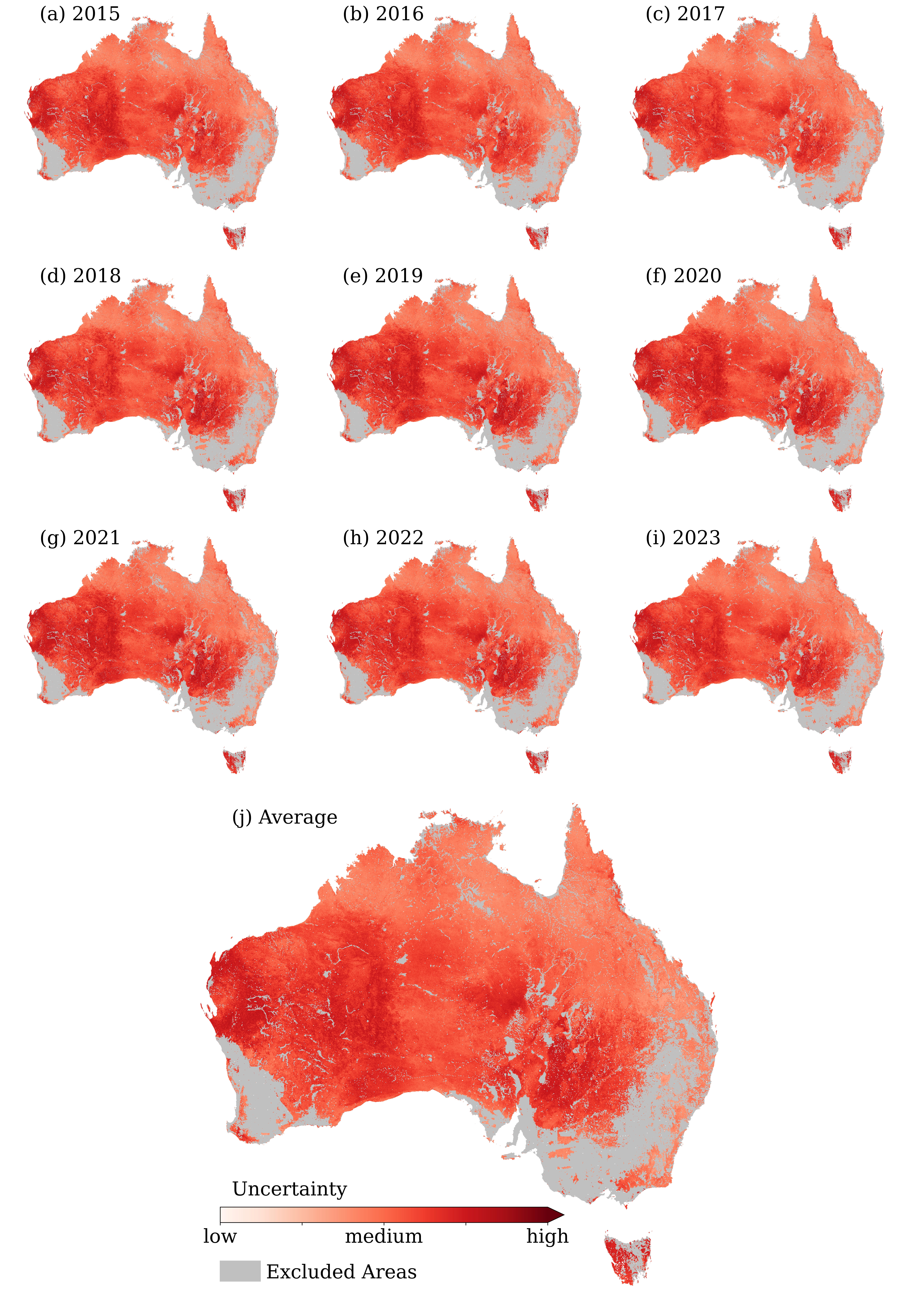}
\end{minipage}
\begin{minipage}{.3\textwidth}
\vspace{-11cm}
\caption{The uncertainty maps for plant species richness predictions from 2015 to 2023 (a--i), and the average uncertainty map across these years (j). The uncertainty values were calculated with the Monte Carlo Dropout approach \cite{gal2016dropout} under a dropout rate of 0.5 for the proposed Spatioformer model.\label{fig:uncertainty}}
\end{minipage}
\end{figure*}

Fig. \ref{fig:uncertainty} shows the uncertainty maps for plant species richness predictions from 2015 to 2023 (Fig. \ref{fig:uncertainty}a--f) and the average uncertainty map across these years (Fig. \ref{fig:uncertainty}g). By comparing the uncertainty maps with the distribution of in-situ samples shown in Fig. \ref{fig:location}, it was found that regions of denser sample coverage, such as the coastal regions, tended to have a lower prediction uncertainty than the interior of Australia with sparse or no coverage of in-situ samples.

It was seen from Fig. \ref{fig:uncertainty} that the eastern, southeastern, and northern coastal regions showed a lower uncertainty than the middle and mid-west interior of Australia, as well as the island of Tasmania. Compared with Fig. \ref{fig:location} where the locations of ground richness samples are displayed, the high uncertainty regions shown in Fig. \ref{fig:uncertainty} mostly corresponded to places where fewer survey samples had been collected. Therefore, this uncertainty map could serve as a guidance as to where future field surveys should be focusing on. While conducting field experiments in remote locations could be considerably arduous, samples to be collected in high uncertainty areas would be of high value in helping improve the overall accuracy of plant species richness maps throughout the country.

It is worth noting that several factors could contribute to the uncertainty in richness predictions. Areas with sparse ground sampling, such as the interior of Australia, exhibited higher prediction uncertainties, highlighting the model's sensitivity to the availability of ground-truth data. Moreover, the quality of in-situ samples used for model training may affect the accuracy of the predictions. The model's performance is another critical factor, as its ability to generalise across diverse landscapes would also influence the reliability of the predicted richness values. 

\subsection{Impact of Input Image Size}

The accuracies of plant species richness prediction achieved with different input image sizes from 1 {\texttimes} 1 to 9 {\texttimes} 9 pixels are shown in Table \ref{tab:impact}. It was observed from the table that, when the input image size increased from 1 {\texttimes} 1 to 9 {\texttimes} 9 pixels, improved mapping accuracies were observed, while the improvement became marginal with more pixels being added especially when the input image size was larger than 5 {\texttimes} 5 pixels. 

The results in Table \ref{tab:impact} indicated that, the spatial information, in combination with the spectral features provided by the imagery, could be beneficial in modelling the relationship between satellite observations and on-ground richness values. Considering that plants usually live as a community where each plant interacts with its spatial neighbours and surrounding habitat conditions in a close relationship, additional spatial information provided by the input image could better inform the model of landscape features. Supplying more pixels as input allowed the model to explore the spectral heterogeneity or variability among these pixels (\emph{i.e.}, the spatial variability in remotely sensed signals), considering that spectral diversity relates closely to biodiversity according to the Spectral Variation Hypothesis proposed by \cite{palmer2002quantitative}. A wider spatial range of the input image also allowed information on spatial heterogeneity and texture to be incorporated into the model. 

\begin{table*}[tb!]
\centering
\caption{Accuracies of the proposed Spatioformer model in plant species richness mapping with input image size varying from 1 {\texttimes} 1 to 9 {\texttimes} 9 pixels. Evaluation metrics include coefficient of correlation (\textit{r}), coefficient of determination (\textit{r}\textsuperscript{2}), Mean Absolute Error (MAE), Relative Absolute Error (RAE), Mean Squared Error (MSE), Relative Squared Error (RSE), and Root Mean Squared Error (RMSE) between model-predicted and ground-truth species richness.\label{tab:impact}}
\renewcommand{\arraystretch}{1.3} 
\begin{tabular}{p{2.5cm}p{2.5cm}p{2.5cm}p{2.5cm}p{2.5cm}p{2.5cm}}
\hline
Metric      & \multicolumn{5}{l}{Input Image Size}                                         \\
            & 1 × 1 Pixel & 3 × 3 Pixels & 5 × 5 Pixels  & 7 × 7 Pixels  & 9 × 9 Pixels    \\ \hline
\textit{r}  & 0.66        & 0.71         & 0.74 & 0.76 & 0.77   \\
\textit{r}\textsuperscript{2} & 0.44        & 0.50         & 0.55 & 0.58 & 0.59 \\
MAE         & 8.62        & 8.27         & 8.04          & 7.93 & 7.83   \\
RAE         & 0.31        & 0.30         & 0.29 & 0.29 & 0.29   \\
MSE         & 165.38      & 123.65       & 111.88        & 107.74       & 105.86 \\
RSE         & 0.13        & 0.12         & 0.12 & 0.11 & 0.11   \\
RMSE        & 12.86       & 11.12        & 10.53         & 10.38         & 10.29  \\ \hline
\end{tabular}
\end{table*}

\subsection{Limitations and Future Work}

In this study, we used Landsat Geomedian products to produce annual maps of plant species richness in Australia. A novel transformer architecture, Spatioformer, which is capable of taking in geo-coordinates, was proposed for the modelling, in order to accommodate the location-dependent relationship between satellite observations and richness measurements. The Spatioformer model outperformed state-of-the-art models in richness prediction accuracy, as demonstrated in Section \ref{ssec:mapping_results}. Nevertheless, there remains potential for enhancing the accuracy further, as discussed below.

This work used remote sensing imagery as data source for plant species richness mapping. As demonstrated in \cite{mokany2022patterns} and \cite{cai2023global}, many environmental variables, such as temperature, precipitation, soil texture, and topographic heterogeneity, are closely related to the spatial pattern of richness, though they are less effective than remote sensing data in capturing temporal changes such as those caused by deforestation, floods, bushfires, and land use intensification. Hence, in future studies, it is worth exploring the combination of environmental variables and remote sensing data to enhance the accuracy of plant species richness mapping.

Landsat's unique long-term record allowed us to cover a large number of historical survey samples, resulting in a decent-sized dataset of richness vs. image pairs for modelling. As shown in \cite{guo2023plant}, both multispectral and hyperspectral satellite data demonstrated a reasonably strong correlation with on-ground plant species richness, with hyperspectral data showing better performance than multispectral. The deployment of a new generation of hyperspectral satellites, such as DESIS \cite{eckardt2015desis}, PRISMA \cite{pignatti2013prisma}, and EnMAP \cite{guanter2015enmap}, provides an opportunity to map plant species richness from hyperspectral data. However, it's important to note that the limited temporal and spatial coverage of these hyperspectral data may pose a major challenge in composing training datasets, as many historical survey samples would not be able to pair with a satellite observation.

In this study, plant species richness was predicted with spaceborne remote sensing data, which can be acquired at a lower cost than in-situ data. However, we recognise that this method cannot fully replace in-situ sampling. Field surveying is the gold-standard approach to richness measurement and provides ground-truth data for model validation. With remote sensing data as a proxy, our study aimed to extend the value of in-situ samples: We first modelled the relationship between richness samples and satellite observations, and then extrapolated the modelled relationship to locations where in-situ samples are absent. In this way, knowledge in the survey samples is expanded to a wider spatial and temporal range, with satellite observations supplementing in-situ measurements to achieve the spatiotemporal resolutions required for monitoring the richness dynamics.

While the Spatioformer model demonstrates good predictive capabilities, understanding the geographical and ecological mechanisms behind its predictions remains an important area for further investigation. In subsequent research, efforts could be made to improve the interpretability of the model by integrating domain knowledge from ecology and geography. Improved model interpretability would not only help explain the spatial and temporal patterns of plant species richness that are related to environmental gradients, habitat fragmentation, and species aggregation, but also provide insights into the ecological processes that drive those patterns.

It is important to note that the relationship between plant species richness and remote sensing imagery could evolve with time due to the emergence of new influencing factors, such as those introduced by human activities and bushfires. Such factors may not be well represented in historical data that our model has been trained on. In order to generalise the ability of our model under future conditions, it is essential to continuously update it by incorporating more recent data reflecting emerging influences. Potential techniques for model updating with new data, such as incremental learning, transfer learning, and model fine-tuning, can be explored in future studies, to incorporate the influence of emerging factors.

As our study focuses on the Australian continent, the findings may be limited when generalizing to regions with different climatic, ecological, or biogeographical characteristics. A recent study \cite{flores2023australia} revealed that Australia is distinctive in a range of plant traits, though its biotic and abiotic characteristics still share some degree of similarity with those in other regions of the world. Therefore, future research should explore our model's performance across different regions and biomes around the world, through incorporating additional data sources that capture local ecological and environmental dynamics. Such efforts would enhance the generalizability of the model and its applicability for global biodiversity monitoring and conservation efforts.

Future work could also focus on extending the Spatioformer model for a broader range of applications. One potential direction is to adapt the model for other biodiversity metrics beyond plant species richness. For instance, the Spatioformer architecture could be modified to test its capability in predicting plant functional diversity \cite{ma2019inferring} or spectral diversity \cite{williams2021remote}, leveraging its ability to incorporate geolocation context into remote sensing data. Moreover, enhancing the model to integrate more complex environmental variables and considering temporal dynamics from long-term satellite observations could improve its applicability to diverse ecological and conservation challenges. Exploring these potential applications of Spatioformer would further assist biodiversity monitoring and management with large-scale remote sensing data.
\section{Conclusions}
\label{sec:conclusions}

We presented a novel transformer architecture, \textit{Spatioformer}, to map the spatial distribution of plant species richness in Australia from Landsat observations. The results demonstrated the feasibility of applying Spatioformer to a continental-scale in-situ richness dataset (HAVPlot) for compiling large-scale richness maps. These maps derived from spaceborne remote sensing data offered a more comprehensive spatiotemporal representation of richness distribution, as compared with the traditional approach of measuring richness via ground sampling.

The Spatioformer model differs from standard transformer architectures in that it allows encoding of geolocations into remote sensing images whose pixel values are intrinsically associated with geolocation coordinates. With Spatioformer, the location-dependent relationship between plant species richness and spectral features in satellite observations was accommodated in the modelling, with enhanced performance being observed relative to state-of-the-art models. 

With Spatioformer, plant species richness maps over Australia were compiled from Landsat archive for the years from 2015 to 2023. The richness maps produced in this study revealed the spatiotemporal dynamics of plant species richness in Australia. The resultant richness maps provided a useful guidance for developing future conservation strategies and decision-making processes with the aim of preserving plant diversity in Australia. Through quantitative analyses, we identified regions where richness predictions are of high uncertainty. Future in-situ surveys may focus on these areas in order to improve the overall accuracy for richness mapping. We also analysed the impact of input image size on the accuracy of richness prediction, showing that incorporating spatially adjacent pixels into modelling is beneficial in richness modelling with satellite data.

This study provided a basis for further improving the accuracy of plant species mapping. In future studies, it is worthwhile to explore the integration of remote sensing data with environmental variables, the incorporation of seasonal information provided by satellite image time series, and the adoption of spaceborne hyperspectral data for richness modelling.
\section*{Acknowledgements}
\label{sec:acknowledgements}

The data sources of in-situ samples utilised in this study are cited in full in the Supporting information of \cite{mokany2022patterns}, and include: data supplied by Department of Environment and Natural Resources {\copyright} Northern Territory of Australia; Natural Values Atlas (www.naturalvaluesatlas.tas.gov.au), 2022, {\copyright} State of Tasmania; NSW BioNet Flora Survey Data Collection {\copyright} State Government of NSW and Department of Planning, Industry and Environment 2013; Queensland CORVEG Database, ver. 8/3/2019 State of Queensland (Department of Environment and Science, www.des.qld.gov.au/); Victorian Biodiversity Atlas {\copyright} State Government of Victoria (accessed June 2017); NatureMap {\copyright} State Government of Western Australia; NatureMaps {\copyright} State Government of South Australia, Department for Environment and Water; TERN Ausplots, The Univ. of Adelaide (www.adelaide.edu.au), Adelaide, South Australia---supported by the Australian Government through the National Collaborative Research Infrastructure Strategy (NCRIS); Desert Ecology Research Group Plots {\copyright} 2015--2018 Rights owned by the Univ. of Sydney; AusCover {\copyright} 2011--2013 The Univ. of Queensland (Joint Remote Sensing Research Program).

The authors would like to thank the anonymous reviewers for their important and insightful comments for improving this manuscript. Our thanks also go to Dr Robert Woodcock, Mr Geoffrey Squire, and Mr Tisham Dhar at CSIRO for their invaluable advice on large-scale cloud computing, and Dr Simon Ferrier for his guidance on plant biodiversity analysis. We acknowledge the computational resources provided by the Earth Analytics Science and Innovation (EASI) platform and the high-performing computer Bracewell. We also acknowledge the satellite image archive provided by Geoscience Australia's Digital Earth Australia (DEA) program.

\bibliographystyle{IEEEtran}
\bibliography{IEEEabrv, ref}

\begin{IEEEbiography}[{\includegraphics[width=1in,height=1.25in,clip,keepaspectratio]{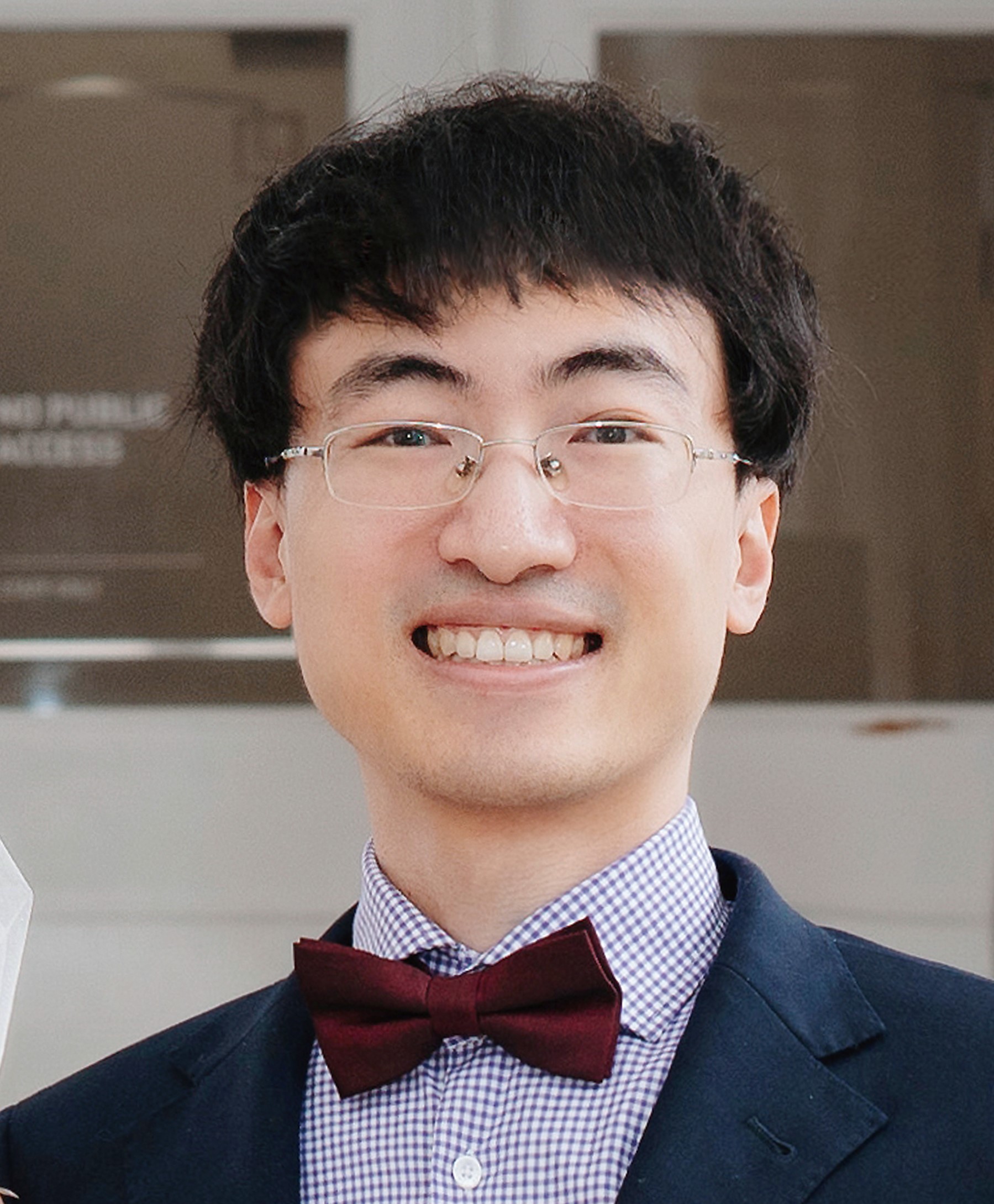}}]{Yiqing Guo}
received the PhD degree from The University of New South Wales (UNSW), Canberra, Australia, in 2019. He is currently a Research Scientist with the Commonwealth Scientific and Industrial Research Organisation (CSIRO), Canberra, Australia. He served as the Inaugural Chair of the IEEE Geoscience and Remote Sensing Society (GRSS) UNSW Canberra Student Chapter from 2018 to 2019. Since 2024, he has been serving as the Chair of the IEEE GRSS Australian Capital Territory and New South Wales Joint Chapter. His research interests include remote sensing and machine learning, and their applications to environmental problems.
\end{IEEEbiography}


\begin{IEEEbiography}[{\includegraphics[width=1in,height=1.25in,clip,keepaspectratio]{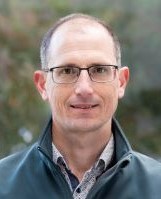}}]{Karel Mokany} received the Ph.D degree from The Australian National University, Canberra, ACT Australia, in 2008. He is currently a Principle Research Scientist with CSIRO, Australia. His research interests include biodiversity modelling, biodiversity assessment and ecology.
\end{IEEEbiography}


\begin{IEEEbiography}[{\includegraphics[width=1in,height=1.25in,clip,keepaspectratio]{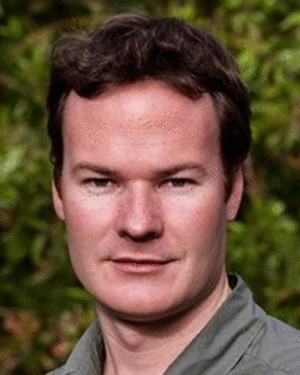}}]{Shaun R. Levick}
received the Ph.D. degree in landscape ecology from the University of the Witwatersrand, Johannesburg, South Africa, in 2008.

He is currently a Principal Research Scientist with the Commonwealth Scientific and Industrial Research Organisation (CSIRO), Darwin, NT, Australia, Australia's national science agency. He specializes in the development and application of remote sensing technologies to address environmental challenges. His research interests include savanna landscapes, particularly, the improvement of sustainable land management and biodiversity conservation.
\end{IEEEbiography}


\begin{IEEEbiography}[{\includegraphics[width=1in,height=1.25in,clip,keepaspectratio]{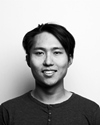}}]{Jinyan Yang}
received B.Sci. degree from East China Normal University, Shanghai, China, in 2010 and the PhD degree from Western Sydney University, Penrith, Australia in 2019. 

He is currently an Early Research Career Fellow in the Commonwealth Scientific and Industrial Research Organisation. His research interests include trait base vegetation modelling, remote sensing, and model data synthesis. 
\end{IEEEbiography}


\begin{IEEEbiography}[{\includegraphics[width=1in,height=1.25in,clip,keepaspectratio]{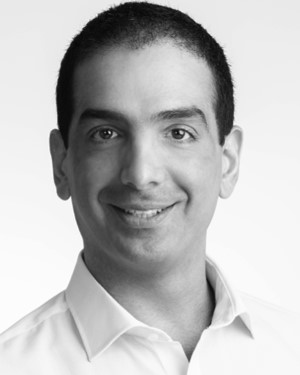}}]{Peyman Moghadam}
is a Principal Research Scientist with CSIRO, DATA61, Brisbane, QLD, Australia, and a Professor (Adjunct) with the Queensland University of Technology (QUT), Brisbane. He leads the Embodied AI Research Cluster, CSIRO, DATA61, working at the intersection of robotics and machine learning. He is also the Spatiotemporal AI Portfolio Leader at CSIRO’s Machine Learning and Artificial Intelligence (MLAI) Future Science Platform and oversees research and development of MLAI methods for scientific discovery in spatiotemporal data streams. In 2022, he was a Visiting Professor at ETH Zürich, Zürich, Switzerland. His research interests include embodied AI, robotics, and machine learning. 
\end{IEEEbiography}



\vfill

\end{document}